\def\year{2019}\relax
\documentclass[letterpaper]{article} 
\usepackage{aaai19}  
\usepackage{times}  
\usepackage{helvet}  
\usepackage{courier}  
\usepackage{url}  
\usepackage{graphicx}  

\usepackage{booktabs}
\usepackage{multirow}
\usepackage{textcomp}
\usepackage{amsmath}
\usepackage{subfig}
\usepackage{epstopdf}
\usepackage{hyperref}

\begin{document}
%
\title{Robust Estimation of Similarity Transformation for Visual Object Tracking}
\author{Yang Li$^1$,
  Jianke Zhu$^{1,3}$\thanks{Corresponding author.},
  Steven C.H. Hoi$^2$,
  Wenjie Song$^1$,
  Zhefeng Wang$^1$,
  Hantang Liu$^1$\\
$^1$Colleage of Computer Science and Technology, Zhejiang University, Hangzhou, China\\
$^2$School of Information Systems, Singapore Management University, Singapore\\
$^3$Alibaba-Zhejiang University Joint Research Institute of Frontier Technologies\\
{\tt\small \{liyang89,jkzhu,seeyou,wangzhefeng,liuhantang\}@zju.edu.cn, chhoi@smu.edu.sg}
}
\maketitle

\def\etal{et.al.}

\begin{abstract}
Most of existing correlation filter-based tracking approaches only estimate simple axis-aligned bounding boxes, and very few of them is capable of recovering the underlying similarity transformation.
  To tackle this challenging problem, in this paper, we propose a new correlation filter-based tracker with a novel robust estimation of similarity transformation on the large displacements.
  In order to efficiently search in such a large 4-DoF space in real-time, we formulate the problem into two 2-DoF  sub-problems and apply an efficient Block Coordinates Descent solver to optimize the estimation result.
  Specifically, we employ an efficient phase correlation scheme to deal with both scale and rotation changes simultaneously in log-polar coordinates.
  Moreover, a variant of correlation filter is used to predict the translational motion individually.
  Our experimental results demonstrate that the proposed tracker achieves very promising prediction performance compared with the state-of-the-art visual object tracking methods while still retaining the advantages of high efficiency and simplicity in conventional correlation filter-based tracking methods.
\end{abstract}

\section{Introduction}\label{sec:introduction}

Visual object tracking is one of the fundamental problems in computer vision with a variety of real-world applications, such as video surveillance and robotics.
Although having achieved substantial progress during past decade, it is still difficult to deal with the challenging unconstraint environmental variations, such as illumination changes, partial occlusions, motion blur, fast motion and scale variations.


\def\fight{0.112\textheight}
\def\fig1line{0.41}
\begin{figure}[!t]
  \centering
    \includegraphics[width=\fig1line\linewidth]{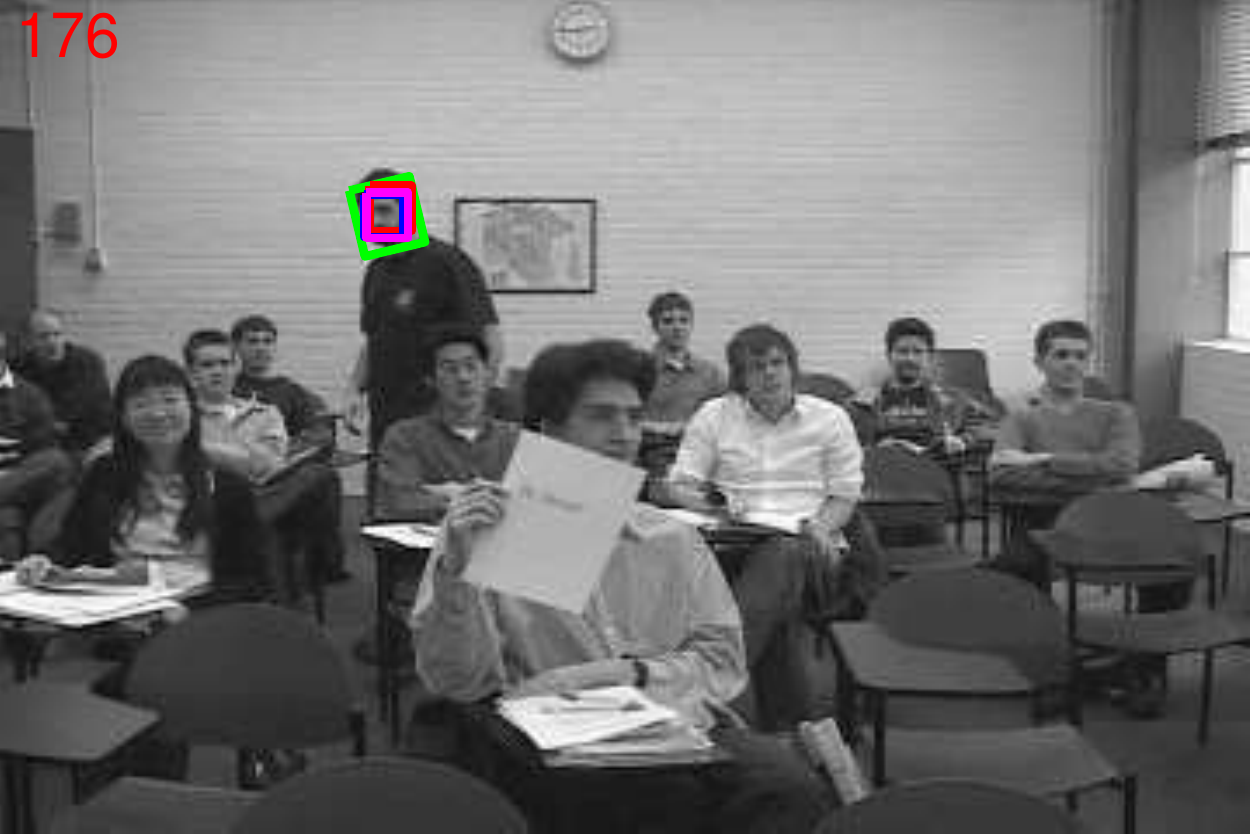}
    \includegraphics[width=\fig1line\linewidth]{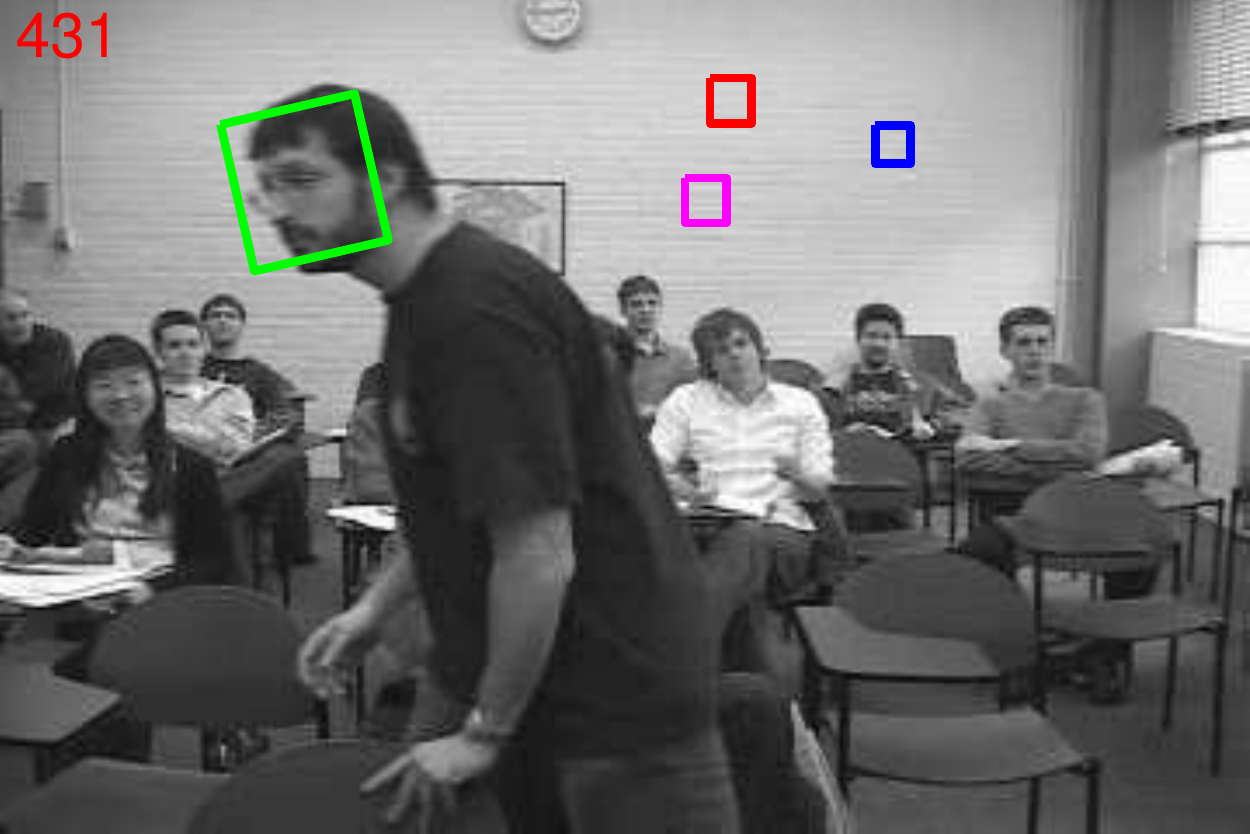}
    \includegraphics[width=\fig1line\linewidth]{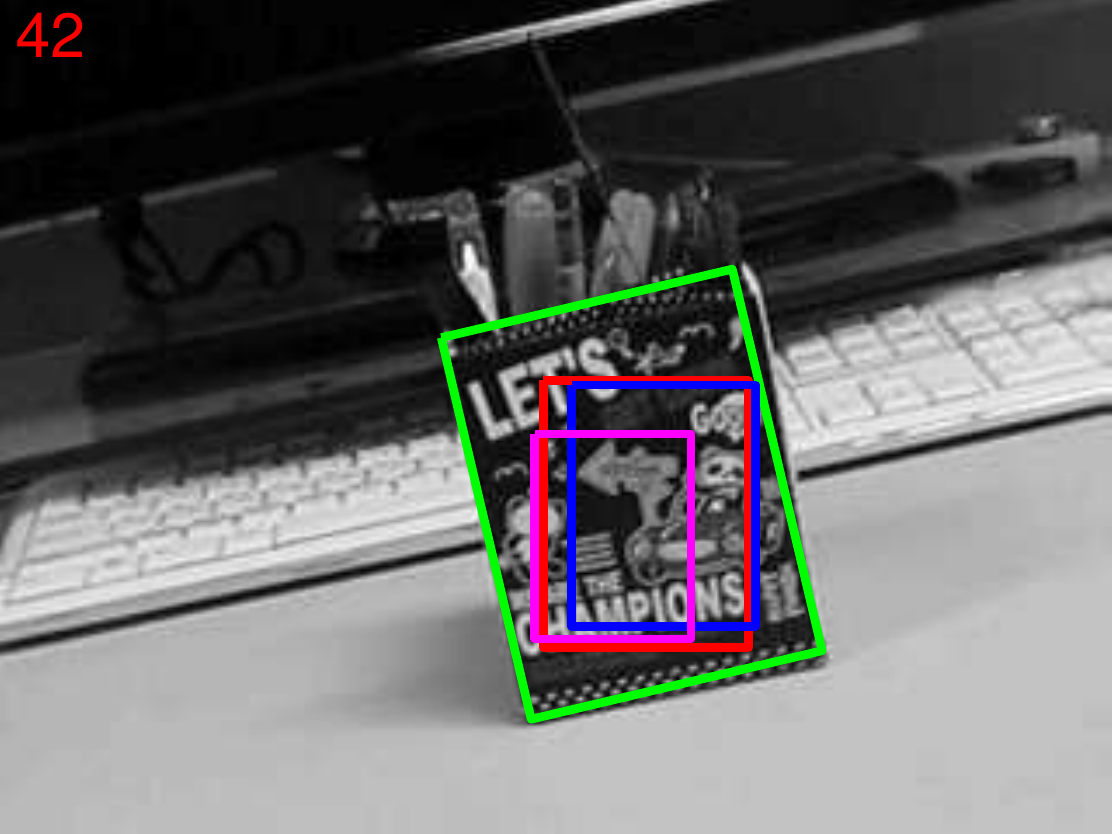}
    \includegraphics[width=\fig1line\linewidth]{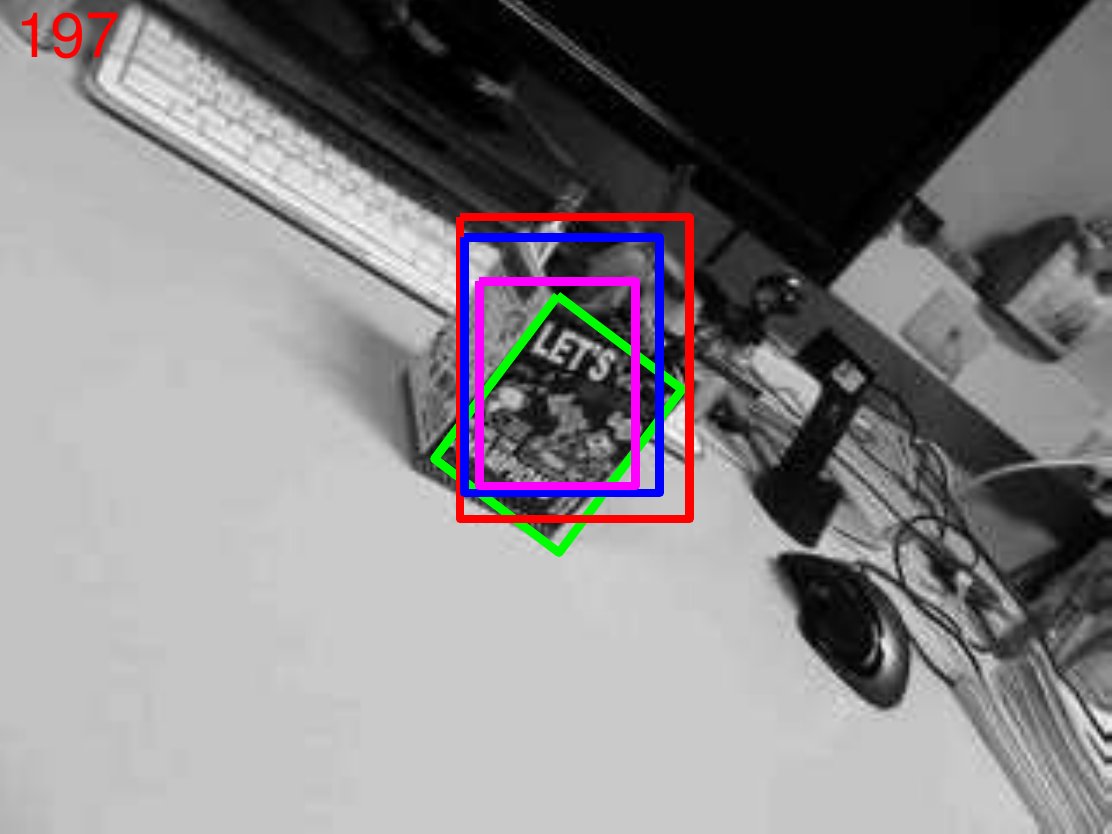}
   
    \includegraphics[width=0.8\linewidth]{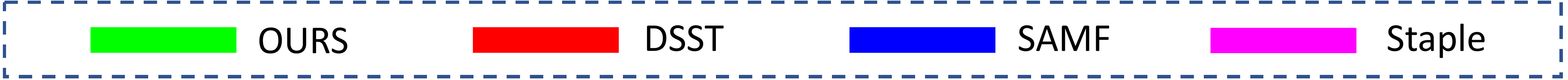}
  \caption {The similarity geometric transformation representation achieves more accurate and robust tracking results. } 
  \label{fig:top}
\end{figure}


\begin{table*}[!t]
  \centering
   \scriptsize
  \caption{Comparison with different kinds of trackers.}
    \label{tab:all}

\begin{tabular}{@{}|l|l|c|c|c|c|c|c|l|@{}}
\toprule
Type & Trackers & Sample Num. &  Scale & Rot. & Pretrain & Performance & GPU & FPS \\ \midrule
\multirow{3}{*}{\begin{tabular}[c]{@{}l@{}}Traditional\\Methods \end{tabular}} & Lucas-Kanade based~\cite{Baker2004Lucas} & Depends  & $\surd$ & $\surd$ & $\times$ & Fair & $\times$ & Depends \\
     & Keypoint based~\cite{cmt} & Depends  & $\surd$ & $\surd$ & $\times$ & Fair & $\times$& 1\texttildelow20 \\
     & Particle filter-based~\cite{ross2008incremental,Ji2012apga} & 300\texttildelow600  & $\surd$ & $\surd$ & $\times$ & Fair & $\times$& 1\texttildelow20 \\
  \hline
\multirow{2}{*}{\begin{tabular}[c]{@{}l@{}}Deep\\ Learning\end{tabular}} & MDNet~\cite{nam2015learning} & 250  & $\surd$ & $\times$ & $\surd$ & Excellent & $\surd$& \texttildelow1 \\
     & SiamFC~\cite{BertinettoVHVT16} & 5  & $\surd$ & $\times$ & $\surd$ & Excellent & $\surd$& 15\texttildelow25 \\
  \hline
  \multirow{4}{*}{\begin{tabular}[c]{@{}l@{}}Correlation\\ Filter\end{tabular}} & original CF~\cite{henriques2015high,bolme2010visual} & 1 &$\times$ & $\times$ &  $\times$ & Fair & $\times$ & 300+ \\
     & DSST~\cite{martin2017fdsst},SAMF~\cite{li2014scale} & 7\texttildelow33 & $\surd$ & $\times$ &  $\times$ &Good &  $\times$ &20\texttildelow80 \\
     & Ours & 2\texttildelow8 & $\surd$ & $\surd$ & $\times$ & Excellent & $\times$ & 20\texttildelow30 \\
  \hline
\end{tabular}
\end{table*}

Recently, correlation filter-based methods have attracted continuous research
attention~\cite{mueller2017ca,Ma-ICCV-2015,ma2015cvpr,zhang2017largedisplacement,li2017cfnn} due to its superior performance and robustness in contrast to traditional tracking approaches.
However,
with correlation filters, little attention has been paid on how to efficiently and precisely estimate scale and rotation changes, which are typically represented in a 4-Degree of Freedom (DoF) similarity transformation. 
To deal with scale changes of the conventional correlation filter-based trackers,~\cite{martin2017fdsst} and~\cite{li2014scale} extended the 2-DoF representation of original correlation filter-based methods to 3-DoF space, which handles scale changes in object appearance by introducing a pyramid-like scale sampling ensemble.
Unfortunately, all these methods have to intensively resample the image in order to estimate the geometric transformation, which incurs huge amounts of computational costs.
In addition, their accuracy is limited to the pre-defined dense sampling of the scale pool.
This makes them unable to handle the large displacement that is out of the pre-defined range in the status space.
Thus, none of these methods is guaranteed to the optimum of the scale estimation.
On the other hand, rotation estimation for the correlation filter-based methods has not been fully exploited yet, since it is very easy to drift away from the inaccurate rotation predictions.
This greatly limits their scope of applications in various wide situations. 
Table~\ref{tab:all} summarizes the
properties of several typical trackers.

To address the above limitations, in this paper, we propose a novel visual object tracker to estimate the similarity transformation of the target efficiently and robustly.
Unlike existing correlation filter-based trackers, we formulate the visual object tracking into a status space searching problem in a 4-DoF status space, which gives a more appropriate geometric transformation parameterization for the target. As shown in Fig.~\ref{fig:top}, the representation in similarity transformation describes the object more correctly and helps to track the visual object more accurately.
To yield real-time tracking performance in the 4-DoF space, we propose to tackle the optimization task of estimating the similarity transformation by applying an efficient Block Coordinates Descent (BCD) solver.
Specifically, we employ an efficient phase correlation scheme to deal with both scale and rotation changes simultaneously in log-polar coordinates and utilize a fast variant of correlation filter to predict the translational motion.
This scheme sets our approach free from intensive sampling, and greatly boosts the performance in the 4-DoF space.
More importantly, as BCD searches the entire similarity transformation space, the proposed tracker achieves very accurate prediction performance in large displacement motion while still retaining advantages of the efficiency and simplicity in conventional correlation filter.
Experimental results demonstrate that our approach is robust and accurate for both generic object and planar object tracking.

The main contributions of our work are summarized as follows: 1) a novel framework of similarity transformation estimation which only samples once for correlation filter-based trackers; 2) a joint optimization to ensure the stability in translation and scale-rotation estimation;
3) a new approach for scale and rotation estimation with efficient
implementation which can improve a family of existing correlation filter-based
trackers~(our implementation is available at \url{https://github.com/ihpdep/LDES}).

\section{Related Work}

Traditionally, there are three genres to handle scale and rotation changes.
The most widely used approach is to iteratively search in an affine status space with gradient descent-based method~\cite{Baker2004Lucas,wenjie2016rpca}.
However, they are easy to get stuck at local optima, which are not robust for large displacements.
Trackers based on particle filter~\cite{ross2008incremental,Ji2012apga,zhang2017mcpf,li2015reliable} search the status space stochastically by observing the samples, which are employed to estimate the global optima in the status space.
Their results are highly related to the motion model that controls the distribution of the 6-DoF transformation.
This makes the tracker perform inconsistently in different situations.
Another choice is to take advantage of keypoint matching to predict the geometric transformation~\cite{cmt,2010ferns}.
These keypoint-based trackers first detect feature points, and then find the matched points in the following frames.
Naturally, they can handle any kind of transformations with the matched feature points.
Due to the lack of global information on the whole target, these trackers cannot
effectively handle the general objects~\cite{kristan2015visual}.

Our proposed method is highly related to correlation filter-based trackers~\cite{henriques2015high,bolme2010visual}.
~\cite{martin2017fdsst} and~\cite{li2014scale} extend the original correlation filter to adapt the scale changes in the sequences.
~\cite{bertinetto2015staple} combines color information with correlation filter method in order to build a robust and efficient tracker.
Later,~\cite{martin2015srdcf} and~\cite{galoogahi2017bacf} decouple the relationship between the size of filter and searching range.
These approaches enable the correlation filter-based methods to have larger searching range while maintaining a relative compact presentation of the learned filters.
~\cite{mueller2017ca} learns the filter with the additional negative samples to enhance the robustness. 
Note that all these approaches emphasize on the efficacy issue, which employs either DSST or SAMF to deal with the scale changes.
However, 
these methods cannot deal with rotation changes.



Fourier Mellin image registration and its variants~\cite{gopalan1994rotationInvariant,siavash2005logpolarRegistraion} are also highly related to our proposed approach.
These methods usually convert both the test image and template into log-polar coordinates, in which the relative scale and rotation changes turn into the translational displacement.
~\cite{gopalan1994rotationInvariant} propose a rotation-invariant correlation filter to detect the same object from a god view.
~\cite{siavash2005logpolarRegistraion} propose an image registration method to
recover large-scale similarity in spatial domain.
Recently,~\cite{yan2016logpolar} and~\cite{zhang2015cf-rotation} introduce the log-polar coordinates into correlation filter-based method to estimate the rotation and scale.
Compared with their approaches, we directly employ phase correlation operation in log-polar coordinates. Moreover, an efficient Block Coordinates Descent optimization scheme is proposed to 
deal with large motions with real-time performance.

\begin{figure*}[tbp]
  \centering
  \includegraphics[width=0.86\linewidth]{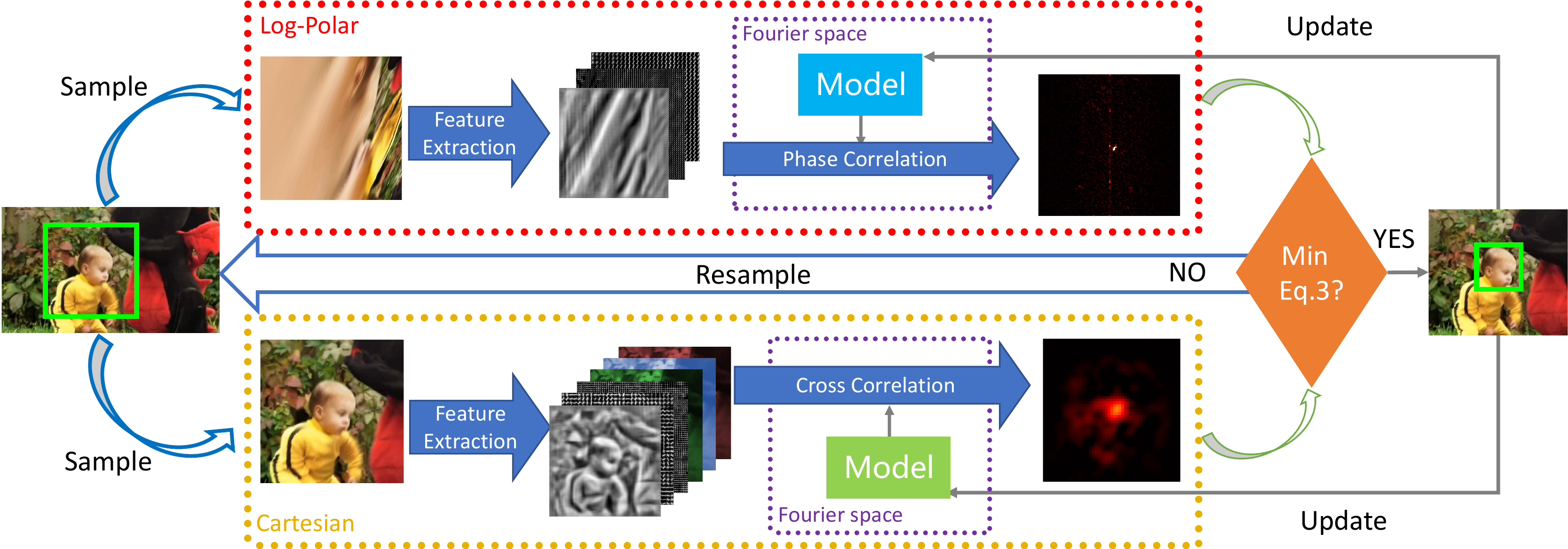}
  \caption{Overview of our proposed approach in estimation of similarity transformation.}
  \label{fig:overview}
  \end{figure*}

\section{Our Approach}
In this paper, we aim to investigate robust visual object tracking techniques to deal with challenging scenarios especially when there are large displacements. We propose a novel robust object tracking approach, named ''Large-Displacement tracking vis Estimation of Similarity~(LDES), where the key idea is to enable the tracker with capability in 2D similarity transformation estimation in order to handle large displacement. Figure~\ref{fig:overview} gives an overview of the proposed LDES approach.
In the following, we first formally formulate the problem as an optimization task, and then divide it into two sub-problems, translation estimation and scale-rotation prediction.
We solve the two sub-problems iteratively to achieve a global optimal.

\subsection{Problem Formulation}
Given an image patch $\mathbf{x}_i$ sampled from the $i$-th frame ${I_i}$ in a video sequence, the key idea of our proposed approach is to estimate the
similarity transformation $Sim(2)$ in 2D image space of the tracked traget. To this end, we need to predict a 4-DoF transformation status vector $\mathbf{\tau}_i \in \mathcal{R}^4 $ based on the output of the previous frame. Generally, $\mathbf{\tau}_i$ is obtained by optimizing the following score function: 

\begin{equation}
 \tau_i = \arg \max_{\tau \in Sim(2)} f(\mathcal{W}( {I}_i,\tau); \mathcal{\mathbf{h}}_{i-1}),
  \label{eq:eq1}
\end{equation}
where $f(\cdot)$ is a score function with the model $\mathbf{h}_{i-1}$ learned from the previous frames $I_{1:i-1}$. $\mathcal{W}$
is an image warping function that samples the image ${I}_i$ with respect to the similarity transformation status vector $\tau$. 

The 2D similarity transformation $Sim(2)$ deals with 4-DoF $\{t_x,t_y,\theta,s\}$ motion, where $\{t_x,t_y\}$ denotes the 2D translation. $\theta$ denotes the in-plane rotation angle, and $s$ represents the scale change with respect to the template. Obviously, $Sim(2)$ has a quite large searching space, which is especially challenging for real-time applications. A typical remedy is to make use of effective sampling techniques to greatly reduce the searching space~\cite{particlefilter}.

Since the tracking model $\mathbf{h}_{i-1}$ is learned from the previous frame, which is kept constant during the prediction. The score function $f$ is only related to the status vector $\tau$. We abuse the notation for simplicity:
\begin{equation}
  f_i(\tau) = f(\mathcal{W}( {I}_i,\tau);\mathbf{h}_{i-1}).
\end{equation}

Typically, most of the conventional correlation filter-based methods only take into account of in-plane translation with 2-DoF, where the score function $f_i$ can be calculated completely and efficiently by taking advantage of Convolution Theorem. To search the 4-DoF similarity space, the total number of candidate status exponentially increases. 

Although Eq.~\ref{eq:eq1} is usually non-convex, the optimal translation is near to the one in the previous frame in object tracking
scenarios. Thus, we assume that the function is convex and smooth in the nearby region, and split the similarity transformation $Sim(2)$ into two blocks, $\mathbf{t}=\{t_x,t_y\}$ and $\rho = \{\theta,s\}$, respectively. We propose a score function $f_i(\tau)$, which is the linear combination of three separate parts: 
\begin{equation}
  f_i(\mathbf{\tau};\mathbf{h}_{i-1}) = \eta f_t(\mathbf{t};\mathbf{h_t}) + (1-\eta) f_{\rho}(\mathbf{\rho};\mathbf{h}_{\rho}) + g(\mathbf{t},\mathbf{\rho}),
  \label{eq:score}
\end{equation}
where $\eta$ is an interpolation coefficient. $f_t$ is the translational score function, and $f_{\rho}$ denotes the scale and rotation score function. $g(\mathbf{t},\mathbf{\rho})=exp(|\tau - \tau_{i-1}|_2)^{-1}$ is the motion model which prefers the location nearby the last status. Please note that we omit the subscript $i-1$ of $\mathbf{h_t}$ and $\mathbf{h}_{\rho}$ for simplicity. 

Eq.~\ref{eq:score} is a canonical form which can be solved by the Block Coordinate Descent Methods~\cite{peter2014cd,yu2010cd}. We optimize the following two subproblems alternatively to achieve the global solution:
\begin{equation}
   \arg \max_{\mathbf{t}} g(\mathbf{t},{\rho}^*) + \eta f_t(\mathbf{t}),
  \label{eq:eq3}
\end{equation}
\begin{equation}
  \arg \max_{{\rho}} g(\mathbf{t}^*,{\rho})+ (1-\eta)f_\rho(\rho) ,
  \label{eq:eq4}
\end{equation}
${\rho}^*$ and $\mathbf{t}^*$ denote the local optimal estimation result from previous iteration, which is fixed for the current subproblem. Since $g$ can be calculated easily, the key to solving Eq.~\ref{eq:eq1} in real-time is to find the efficient solvers for the above two subproblems, $f_\rho$ and $f_t$.

\subsection{Translation Estimation by Correlation Filter}

Translation vector $\mathbf{t}$ can be effectively estimated by Discriminative Correlation Filters (DCF)~\cite{henriques2015high,mueller2017ca}. A large part of its success is mainly due to the Fourier trick and translation-equivariance within a certain range, which calculates the $f_t$ in the spatial space exactly. According to the property of DCF, the following equation can be obtained:
\begin{equation}
  f_t(\mathcal{W}({I},\mathbf{t});\mathbf{h_t}) =  \mathcal{W}( f_t({I};\mathbf{h_t}),\mathbf{t}).
  \label{eq:eq6}
\end{equation}
Since the calculation of $\arg \max_{\mathbf{t}} \mathcal{W}( f_t({I};\mathbf{h_t}),\mathbf{t})$ is unrelated to $\mathcal{W}$, we can directly obtain the transformation vector $\mathbf{t}$ from the response map.
Thus, the overall process is highly efficient. The score function $f_t$ can be obtained by
\begin{equation}
  f_t(\mathbf{z}) =  \mathcal{F}^{-1}\sum_{k}\hat{\mathbf{h_t}}^{(k)}\odot\hat\Phi^{(k)}(\mathbf{z}),
  \label{eq:eq7}
\end{equation}
where $\mathbf{z}$ indicates a large testing patch.
$\mathcal{F}^{-1}$ denotes the inverse Discrete Fourier Transformation operator, $\odot$ is the element-wise multiplication and $\hat{\cdot}$ indicates the Fourier space.
$\mathbf{h_t}^{(k)}$ and $\Phi^{(k)}$ represent the $k$-th channel of the linear model weights and the feature map, respectively.
The whole computational cost is $\mathcal{O}(KN\log N)$, where $K$ is the channel number and $N$ is the dimension of the patch $\mathbf{z}$. 

To this end, we need to learn a model $\mathbf{h_t}$ in the process. Note that any quick learning method can be used. Without loss of generality, we briefly review a simple correlation filter learning approach~\cite{bolme2010visual} as follows:
\begin{equation}
  \begin{aligned}
      {{{\left\| \sum\limits_k {\Phi^{(k)}(\mathbf{x}) \star {\mathbf{{h}_t}^{(k)}} - {\mathbf{y}}} \right\|}_2^2} + \lambda_1{{\left\| \mathbf{h_t} \right\|}_2^2}}, 
  \label{eq:eq8}
  \end{aligned}
\end{equation}
where $\star$ indicates the correlation operator and $\lambda_1$ is the regularization filters. $\mathbf{y}$ is the desired output, which is typically a Gaussian-like map with maximum value of one. 
 According to Parseval's theorem, the formulation can be calculated without correlation operation.
By stalling each channel and vectorizing the matrix, Eq.~\ref{eq:eq8} can be reformulated as a normal ridge regression without correlation operation. Thus, the solution to
Eq.~\ref{eq:eq8} can expressed as follows:
\begin{equation}
  \mathbf{\hat{h}_t} = (\mathbf{\hat{X}}^T\mathbf{\hat{X}} + \lambda_1\mathbf{I})^{-1}\mathbf{\hat{X}}^T\mathbf{\hat{y}},
\end{equation}
where $\mathbf{\hat{X}} = [\mathbf{diag}({\hat\Phi^{(1)}(\mathbf{x})})^T,...,\mathbf{diag}({\hat\Phi^{(K)}(\mathbf{x})})^T]$
%
and
$\mathbf{\hat{h}_t} =  [\mathbf{\hat{h}_t}^{(1)T},...,\mathbf{\hat{h}_t}^{(K)T}]^T$.
In this form, we need to solve a $KD \times KD $ linear system, where $D$ is the dimension of testing patch $\mathbf{x}$.

\def\stopfig{0.146} 
\begin{figure*}[tbp]
  \centering
  \includegraphics[width=\stopfig\linewidth]{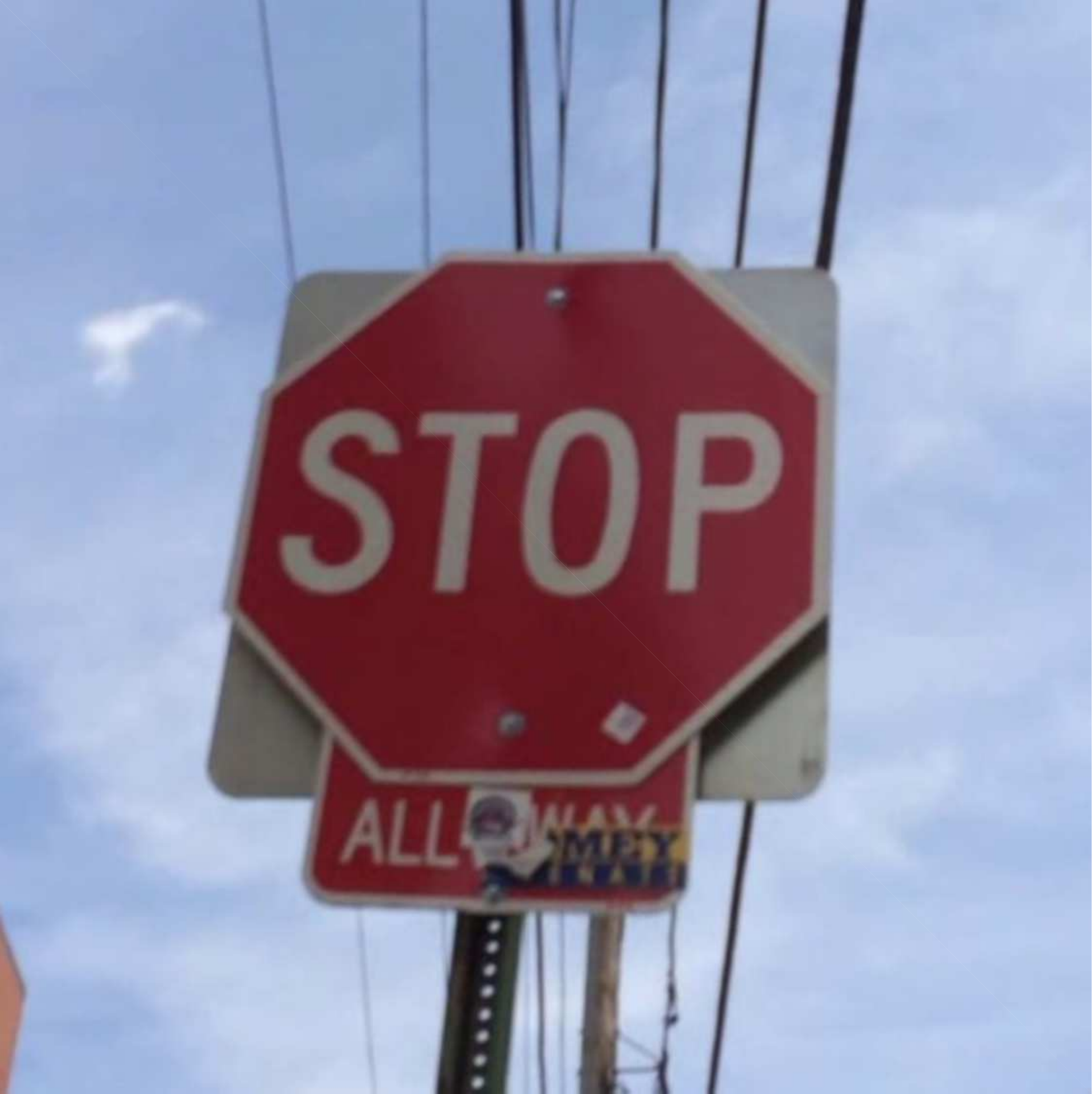}
  \includegraphics[width=\stopfig\linewidth]{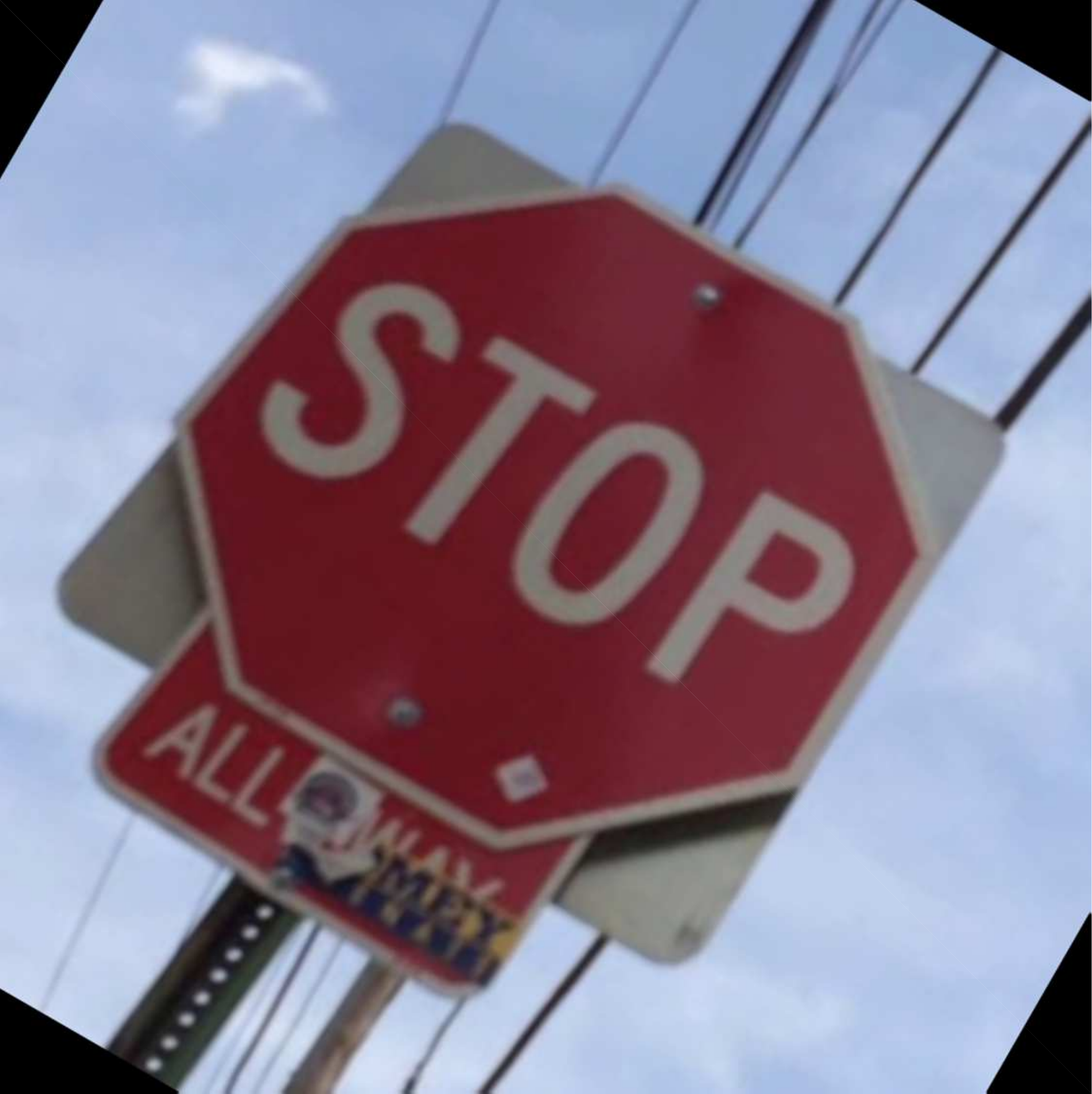}
  \includegraphics[width=\stopfig\linewidth]{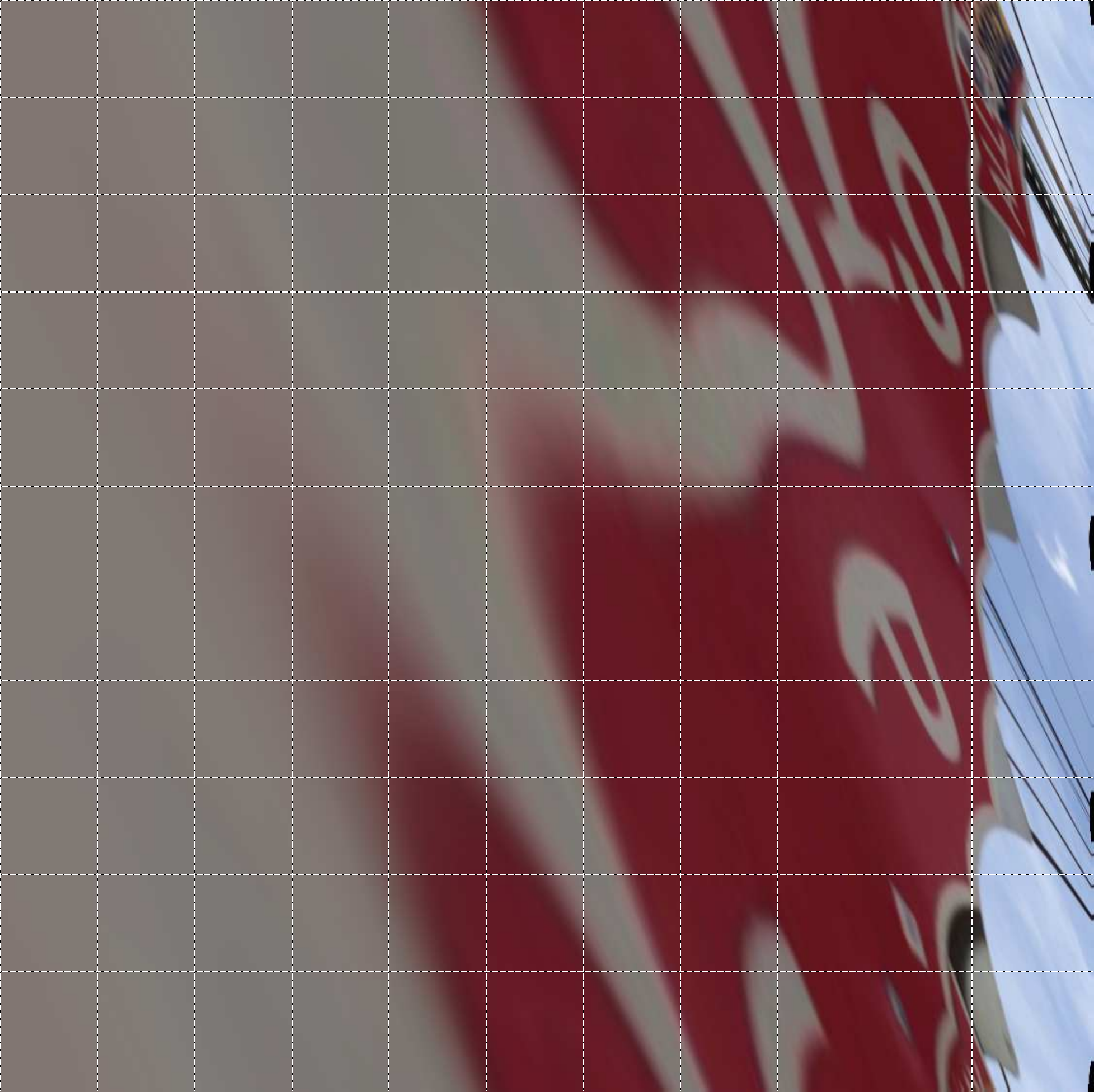}
  \includegraphics[width=\stopfig\linewidth]{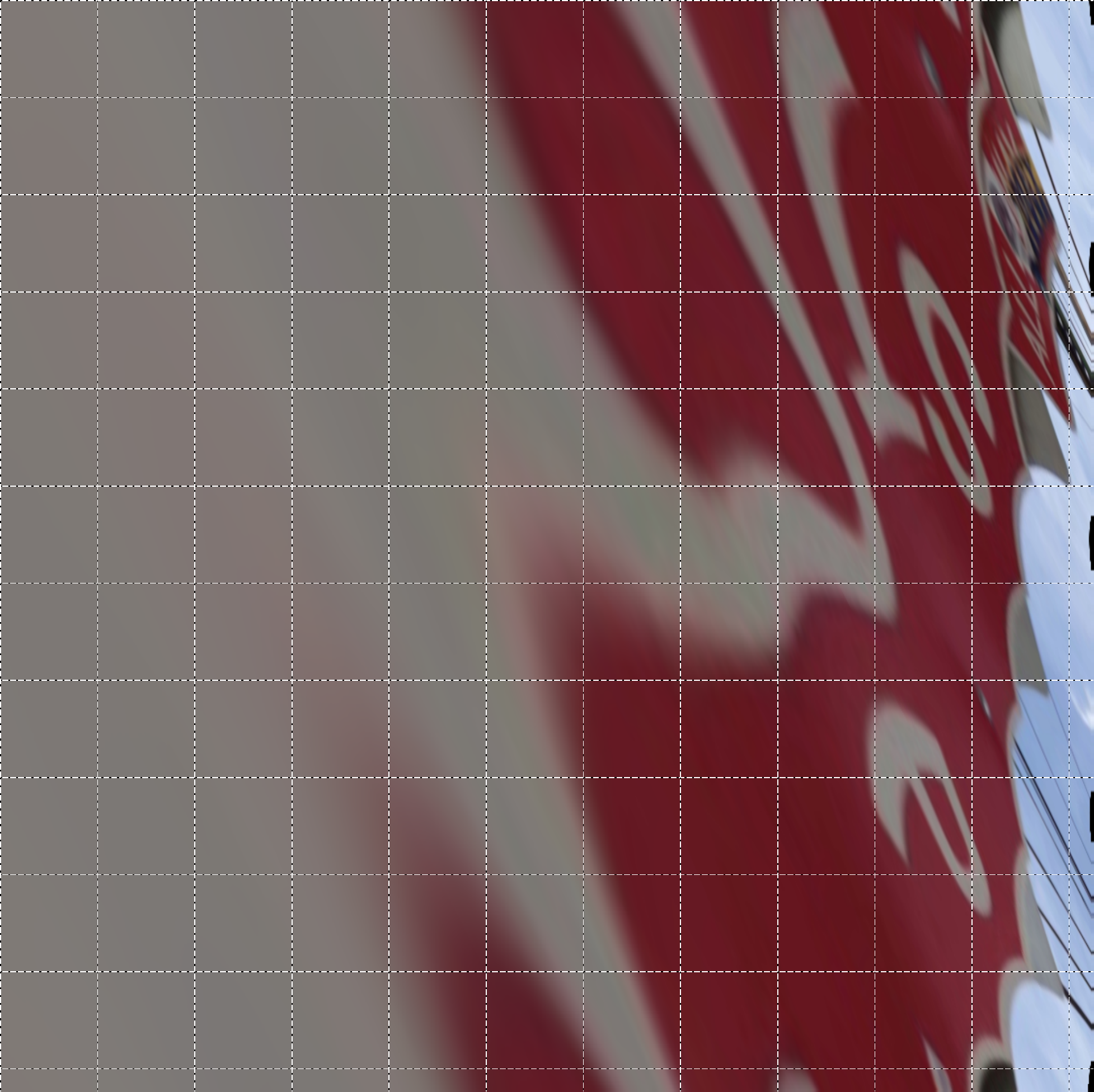}
  \includegraphics[width=\stopfig\linewidth]{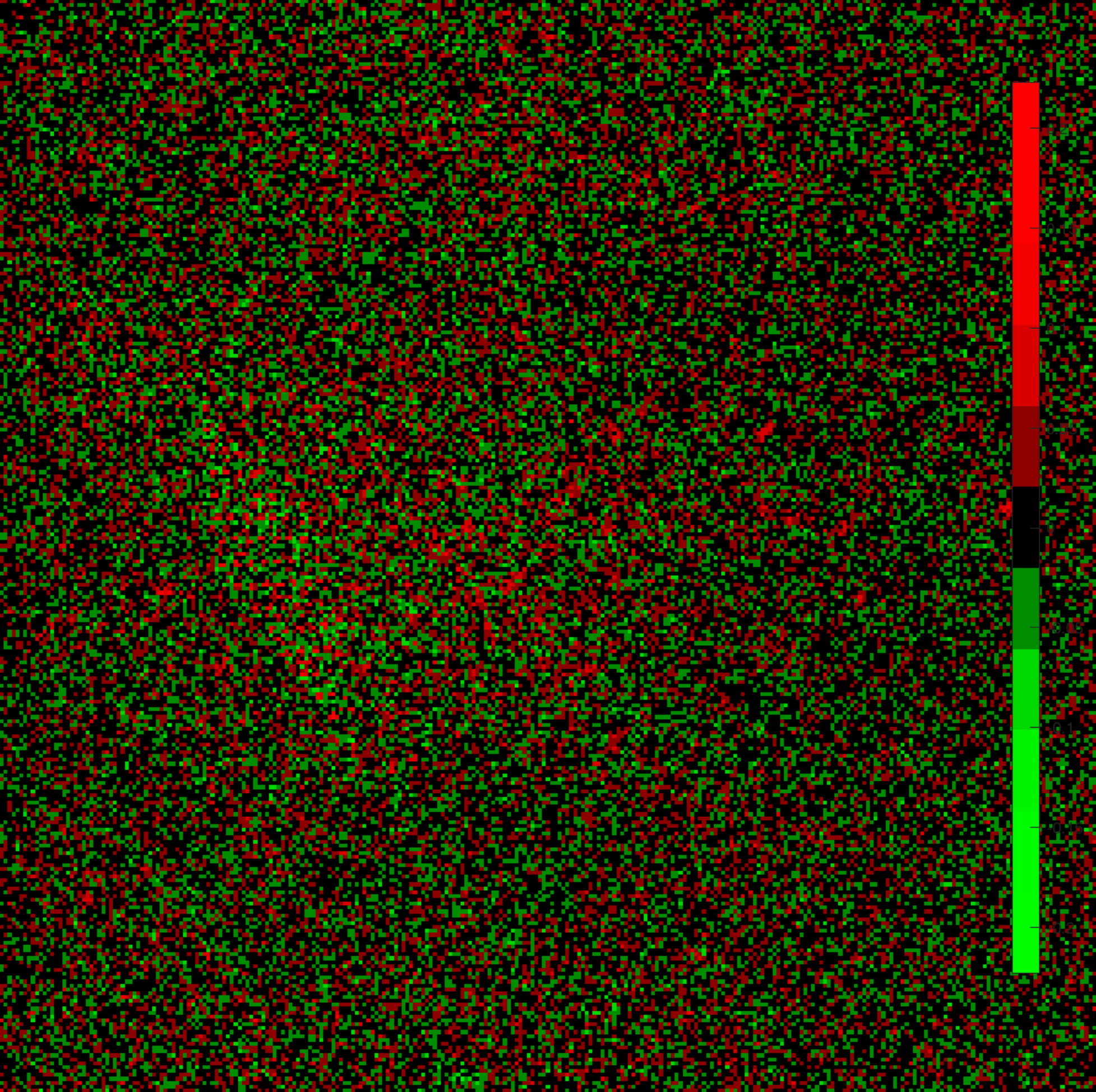}
  \includegraphics[width=\stopfig\linewidth]{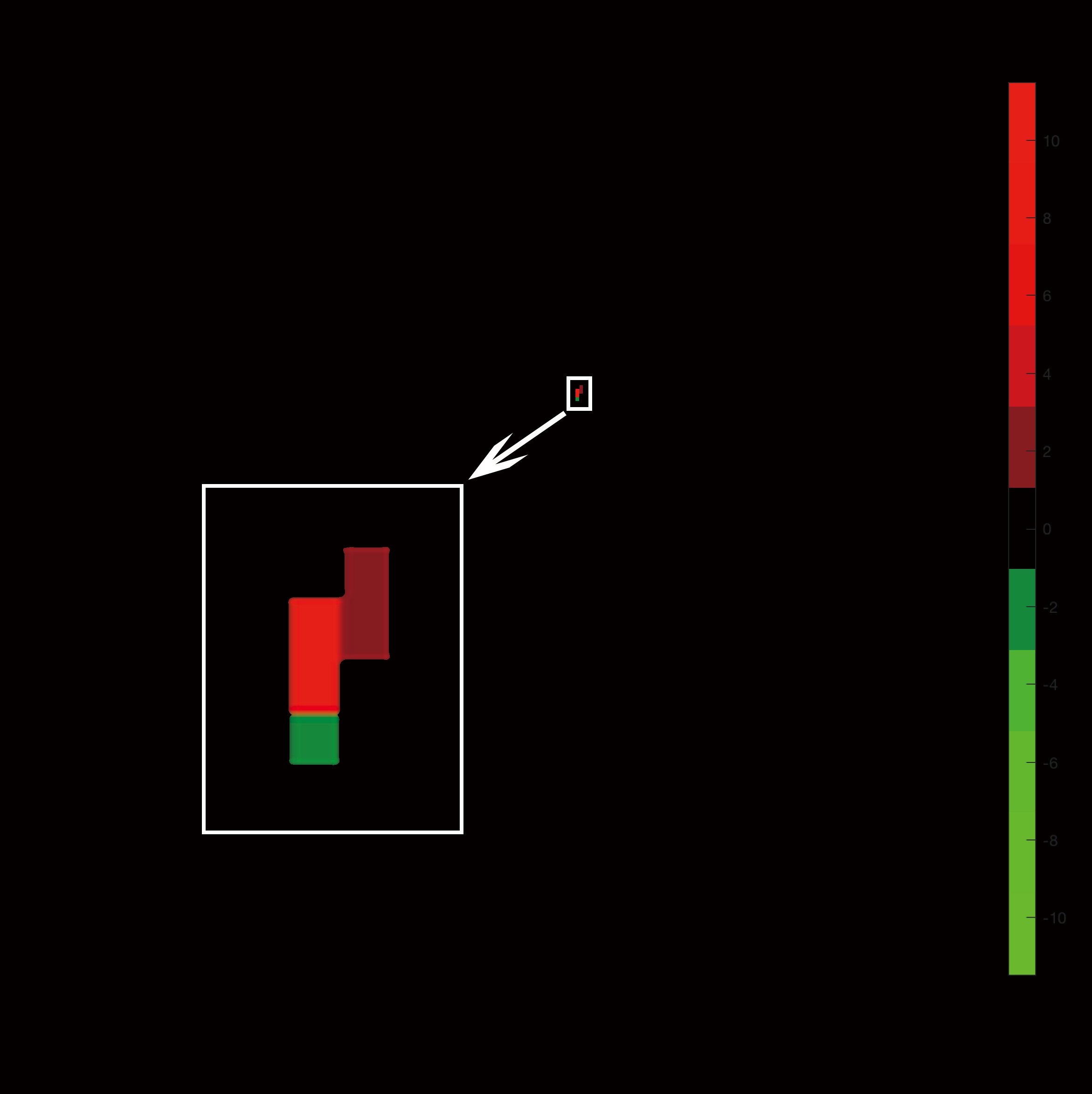}
  
    \caption{ The 3rd and 4th charts are corresponding Log-polar coordinates of the 1st and 2nd images. 2nd image is a $30^{\circ}$ rotation and 1.2 times scale version of the first image.
    The last two charts are the phase correlation response maps. In log-polar coordinates, the response is a peak while it is noisy in Cartesian coordinates.}
     \label{fig:log}
 \end{figure*}

To solve our sub-problem efficiently, we assume that every channel is independent. Thus, by applying Parseval's theorem, the whole system can be simplified as element-wise
operation. The final solution can be derived as below:

\begin{equation}
  \begin{aligned}
    \mathbf{\hat{h}_t}^{(k)}
    &= \hat\alpha \odot {\hat\Psi}^{(k)}\\ 
&=(\mathbf{\hat{y}}\odot^{-1} (\sum_k\hat\Phi^{(k)}(\mathbf{{x}})^* \odot \hat\Phi^{(k)}(\mathbf{{x}}) + \lambda)) \odot \hat\Phi^{(k)}(\mathbf{{x}})^* ,
  \label{eq:eq10}
  \end{aligned}
\end{equation}
where
$\alpha$ denotes the parameters in dual space and $\Psi $ indicates the model sample in feature space. 
$\odot^{-1}$ is the element-wise division.
Thus, the solution can be very efficiently 
obtained with a computational cost of $\mathcal{O}(KD)$. With Eq.~\ref{eq:eq10},
the computational cost of Eq.~\ref{eq:eq8} is $\mathcal{O}(KD\log D)$ which is
dominated by the FFT operation. For more details, please refer
to the seminal work~\cite{henriques2015high,multichannel2013mccf,mueller2017ca}.

\subsection{Scale and Rotation in Log-polar Coordinates}
We introduce an efficient method to estimate scale and rotation changes simultaneously in the log-polar coordinates.

\subsubsection{Log-Polar Coordinates}
Suppose an image $I(x,y)$ in the spatial domain, the log-polar coordinates $I'(s,\theta)$ can be
viewed as a non-linear and non-uniform transformation of the original Cartesian
coordinates. Like polar coordinates, the log-polar coordinates needs a
pivot point as the pole and a reference direction as the polar axis in order to expend
the coordinates system. One of the dimension is the angle between the point and
the polar axis. The other is the logarithm of the distance between the point and the pole.

Given the pivot point $(x_0,y_0)$ and the reference direction $\mathbf{r}$
in Cartesian coordinates, the relationship between Cartesian coordinates and
Log-polar coordinates can be formally expressed as follows:
\begin{equation}
  \begin{aligned}
  &s = \log(\sqrt{(x-x_0)^2 + (y-y_0)^2})\\
  &\theta = \cos^{-1}(\frac{<\mathbf{r},(x-x_0,y-y_0)>}{||\mathbf{r}||\sqrt{(x-x_0)^2 + (y-y_0)^2}}).
  \label{eq:eq11}
  \end{aligned}
\end{equation}

 Usually, the polar axis is chosen as the $x$-axis in Cartesian coordinates, where
 $\theta$ can be simplified as $\tan^{-1}(\frac{y-y_0}{x-x_0})$. Suppose two images are related
 purely by rotation $\tilde{\theta}$ and scale $e^{\tilde{s}}$ which can be written as
 $I_0(e^s\cos\theta,e^s\sin\theta) =
 I_1(e^{s+\tilde{s}}\cos(\theta+\tilde{\theta}),e^{s+\tilde{s}}\sin(\theta+\tilde{\theta}))$
 in Cartesian coordinates.
The log-polar coordinates enjoy an appealing merit that the relationship in the above equation can be derived as the following formula in log-polar coordinates:
\begin{equation}
  I_0'(s,\theta) = I_1'(s+\tilde{s},\theta+\tilde{\theta}),
\end{equation}
  where the pure rotation and scale changes in Log-polar coordinates can be viewed as the translational moving along the axis.  
As illustrated in Fig.~\ref{fig:log}, this property naturally can be employed to estimate the scale and rotation changes of the tracked target.

\subsubsection{Scale and Rotation Changes}
By taking advantage of the log-polar coordinates, Eq.~\ref{eq:eq4} can be calculated very
efficiently. Similarly, scale-rotation invariant can be hold as in Eq.~\ref{eq:eq6}. The scale-rotation can be calculated as below:
\begin{equation}
  f_{\rho}(\mathcal{W}(I_i,\rho);\mathbf{h}_\rho) =  \mathcal{W}( f_{\rho}(I_i;\mathbf{h}_\rho),\rho'),
\end{equation}
where $\rho'=\{\theta',s'\}$ is the coordinates of $\rho$ in log-polar space.
$s = e^{s'\log(W/2)/W}$ and $\theta = 2\pi\theta'/H$. $H$ and $W$ is the height and width of the image $I_i$, respectively.
Similar to estimating the translation vector $\mathbf{t}$ by $f_{t}$, the whole space of $f_{\rho}$ can be computed at once through the Fourier trick:
\begin{equation}
  f_{\rho}(\mathbf{z}) =  \mathcal{F}^{-1}\sum_{k}\mathbf{\hat{h}}_{\rho}^{(k)}\odot\hat\Phi^{(k)}(\mathcal{L}(\mathbf{z})),
  \label{eq:eq14}
\end{equation}
where $\mathcal{L}(x)$ is the log-polar transformation function, and $\mathbf{h}_{\rho}$ is a linear
model weights for scale and rotation estimation.  Therefore, the scale and rotation estimation can be obtained very efficiently without any transformation sampling $\mathcal{W}$. Note that the computational cost of Eq.~\ref{eq:eq14} is unrelated to the sample numbers of scale or rotation. This is extremely efficient compared to the previous enumerate methods~\cite{li2014scale,martin2017fdsst}

To obtain the $\mathbf{\hat{h}}_{\rho}$ efficiently, we employ the phase-correlation to conduct the estimation,
\begin{equation}
  \mathbf{\hat{h}}_{\rho} = \hat{\Upsilon}^* \odot^{-1} |\hat{\Upsilon} \odot \hat\Phi(\mathcal{L}(\mathbf{x}))|,
\end{equation}
where $\Upsilon = \sum_{j}\beta_j\Phi(\mathcal{L}(\mathbf{x}_j))$ is the linear combination of
previous feature patch and $|\cdot|$ is the normal operation.
Intuitively, we compute the phase correlation between current frame and the average of previous frames to align the image. 


\subsection{Implementation Details}
In this work, we alternatively optimize Eq.~\ref{eq:eq3} and
Eq.~\ref{eq:eq4} until $f(\mathbf{x})$ does not decrease or reaches the maximal number
of iterations. After the optimization, we update the correlation filter model as 
\begin{equation}
 \hat\Psi_i = (1-\lambda_{\phi})\hat\Psi_{i-1} +\lambda_{\phi}\hat\Phi(\mathbf{{x_i}}),
\end{equation}
where $\lambda_\phi$ is the update rate of the feature data model in Eq.~\ref{eq:eq10}.
The kernel weight in dual space is updated as below:
\begin{equation}
\begin{aligned}
  \hat{\mathbf{\alpha}}_i =& (1-\lambda_\alpha)\hat{\mathbf{\alpha}}_{i-1} \\
  +&\lambda_\alpha (\hat{\mathbf{y}} \odot^{-1} (\sum_k\hat\Phi^{(k)}(\mathbf{{x}}_i)^*\odot \hat\Phi^{(k)}(\mathbf{{x}}_i) + \lambda_1)),
  \end{aligned}
\end{equation}
where $\lambda_\alpha$ is the update of the kernel parameter in dual space of Eq.~\ref{eq:eq10}. Although there exist some theoretical sounding updating schemes~\cite{kiani2015cfwlb,martin2017eco,martin2015srdcf}, the reason we use linear combination is due to its efficiency and the comparable performance.

Meanwhile, we also update the scale and rotation model as a linear combination,
\begin{equation}
  \Upsilon_i = (1-\lambda_w)\Upsilon_{i-1} + \lambda_w\Phi(\mathcal{L}(\mathbf{x}_i)),
\end{equation}
where $\lambda_w$ can be explained as an exponentially weighted average of the model $\beta_j\Phi(\mathcal{L}(\mathbf{x}_j))$.
We update the model upon $\Phi$ instead of $x_i$ because
$\Phi(\sum_i \mathcal{L}(\mathbf{x}_i))$ is not defined.
The logarithm function in log-polar transformation intends to blur the image due to the nonuniform sampling. This will decrease the visual information in the original images.

To alleviate the artificial effects casued by discretization, we interpolate the $f_{t}$ and $f_{\rho}$  with a centroid-based method to obtain sub-pixel level precision.
In addition,
we use different size of $\mathbf{z}$ in testing and $\mathbf{x}$ in training since a larger search range ($N>D$) help to improve the robustness for the solution to sub-problems.
To match the different dimension $N$ and $D$, we pad $\mathbf{h}$ with zero in spatial space.

\def\tbfig{0.23}
\begin{figure*}[!t]
  \centering
  \subfloat[Overall results on OTB-2013]
  {

\includegraphics[width=\tbfig\linewidth]{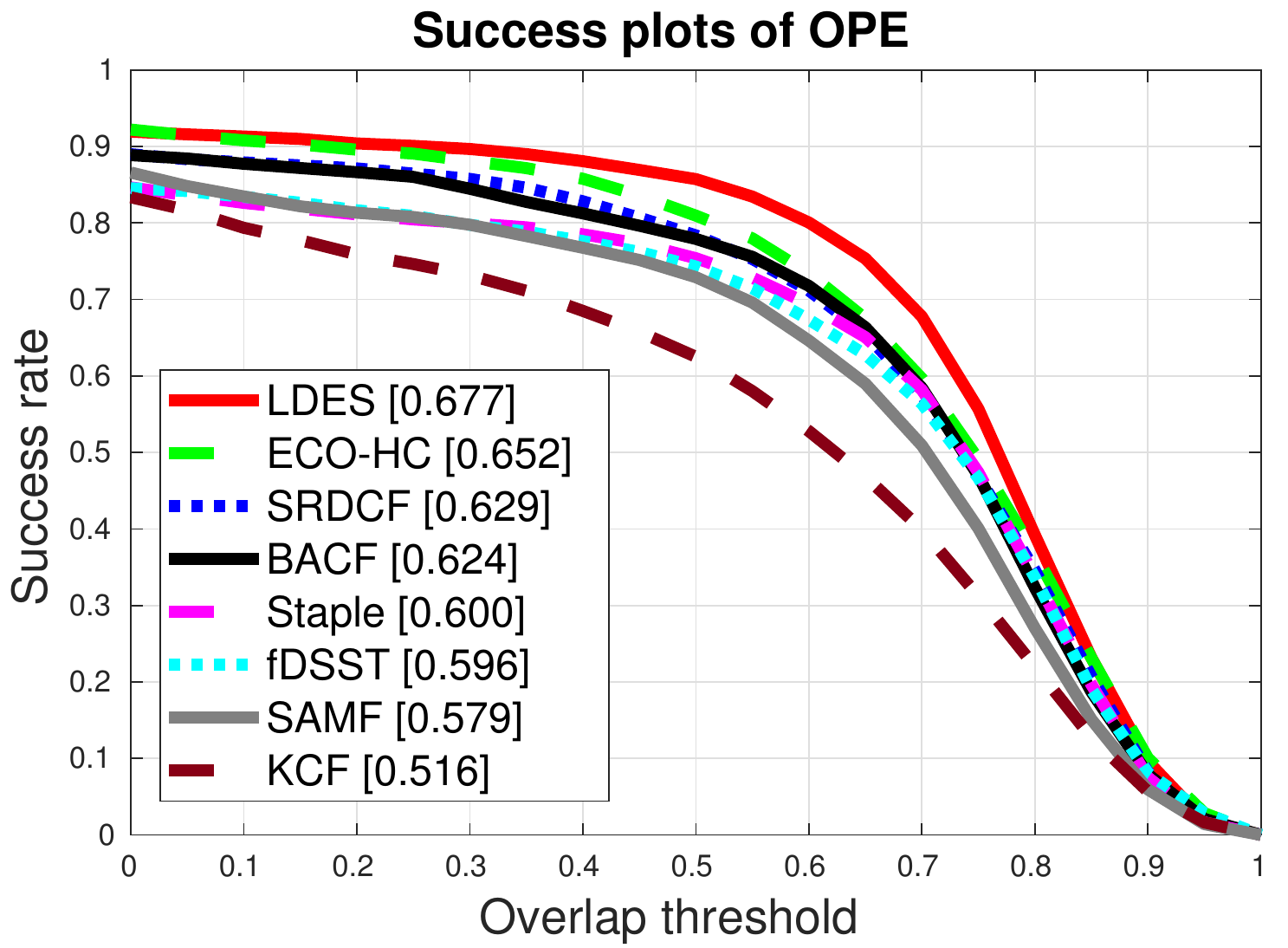}
\includegraphics[width=\tbfig\linewidth]{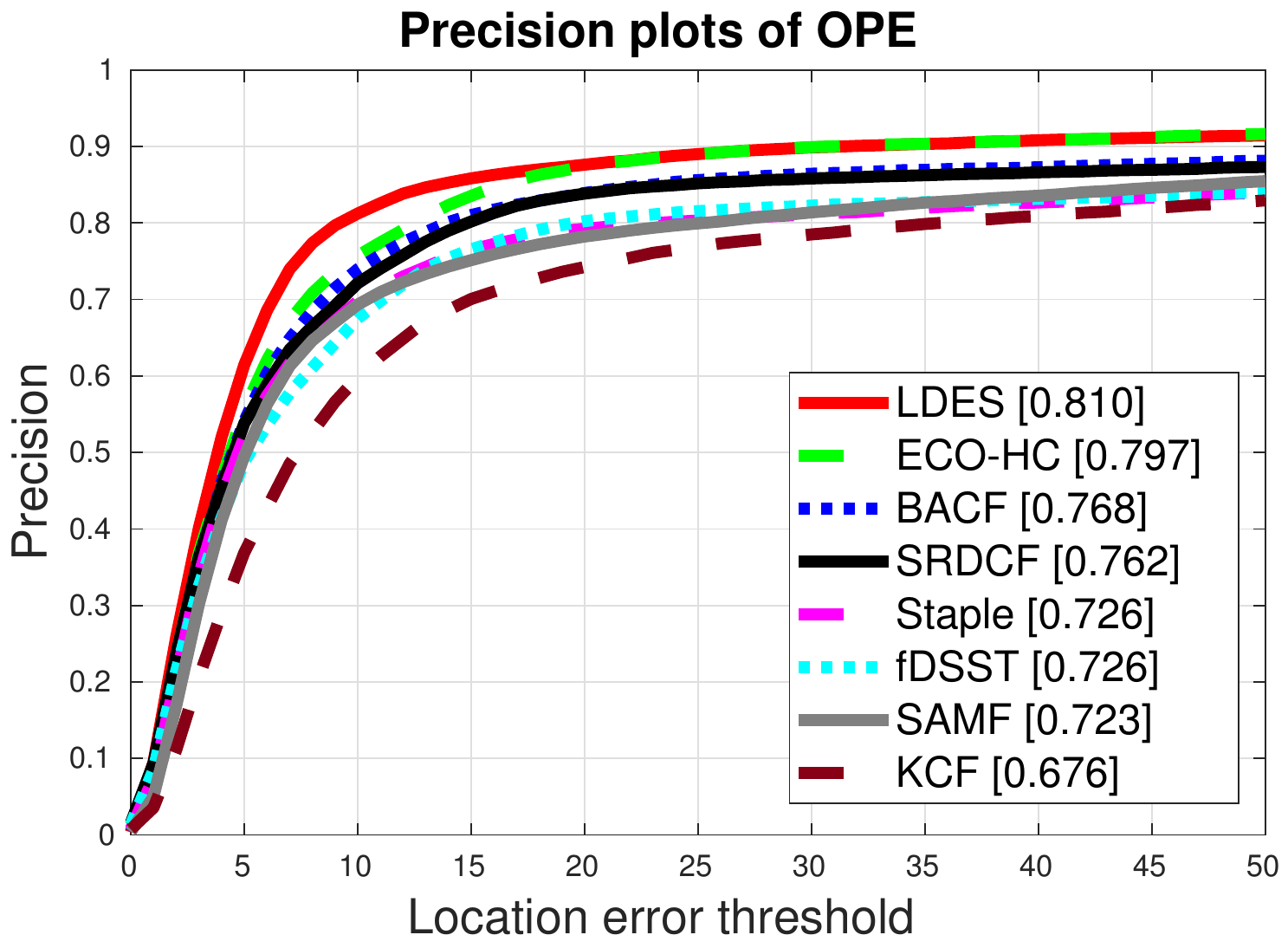}
}
    \subfloat[Overall results on OTB-100]
  {

\includegraphics[width=\tbfig\linewidth]{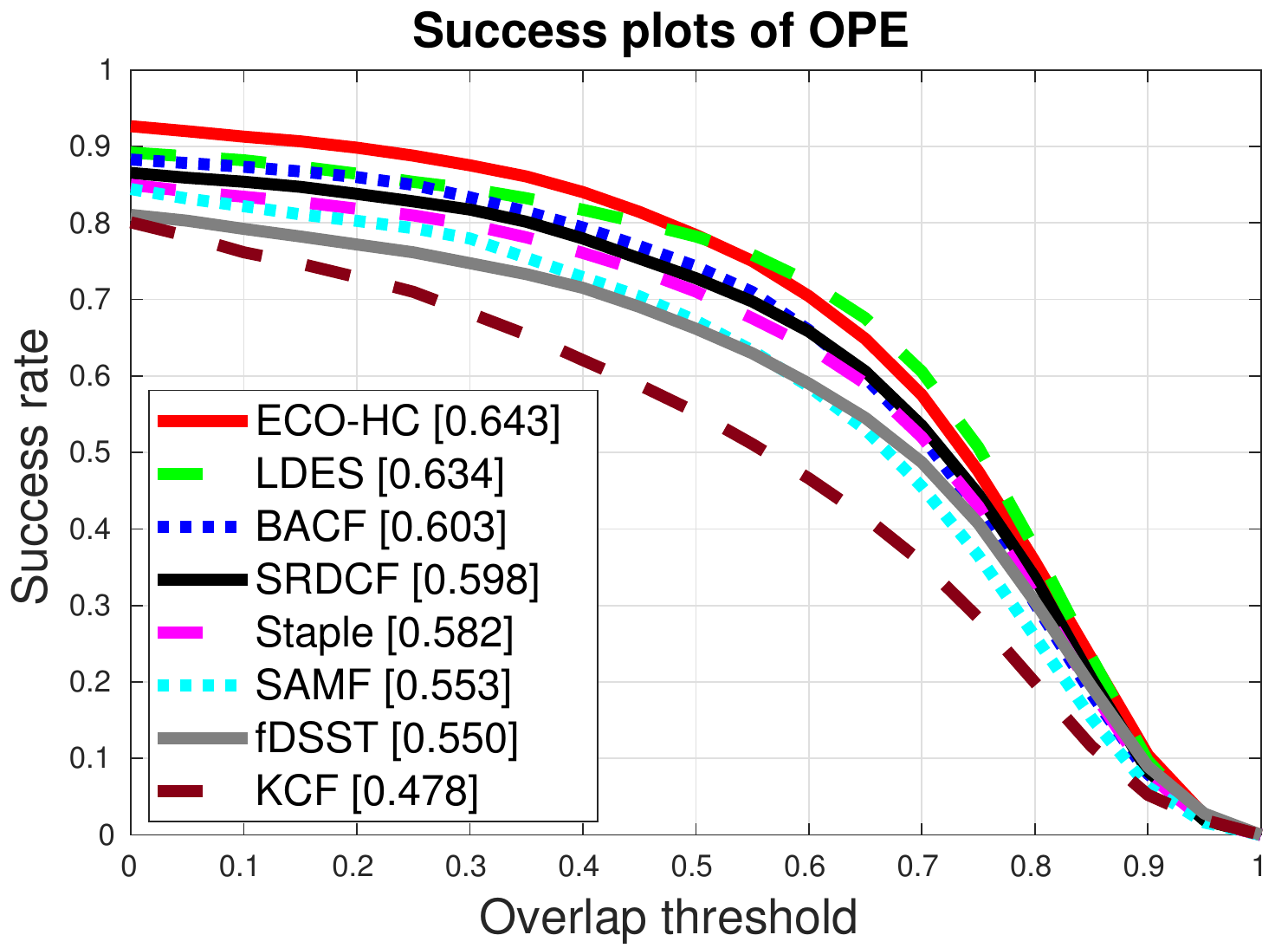}
\includegraphics[width=\tbfig\linewidth]{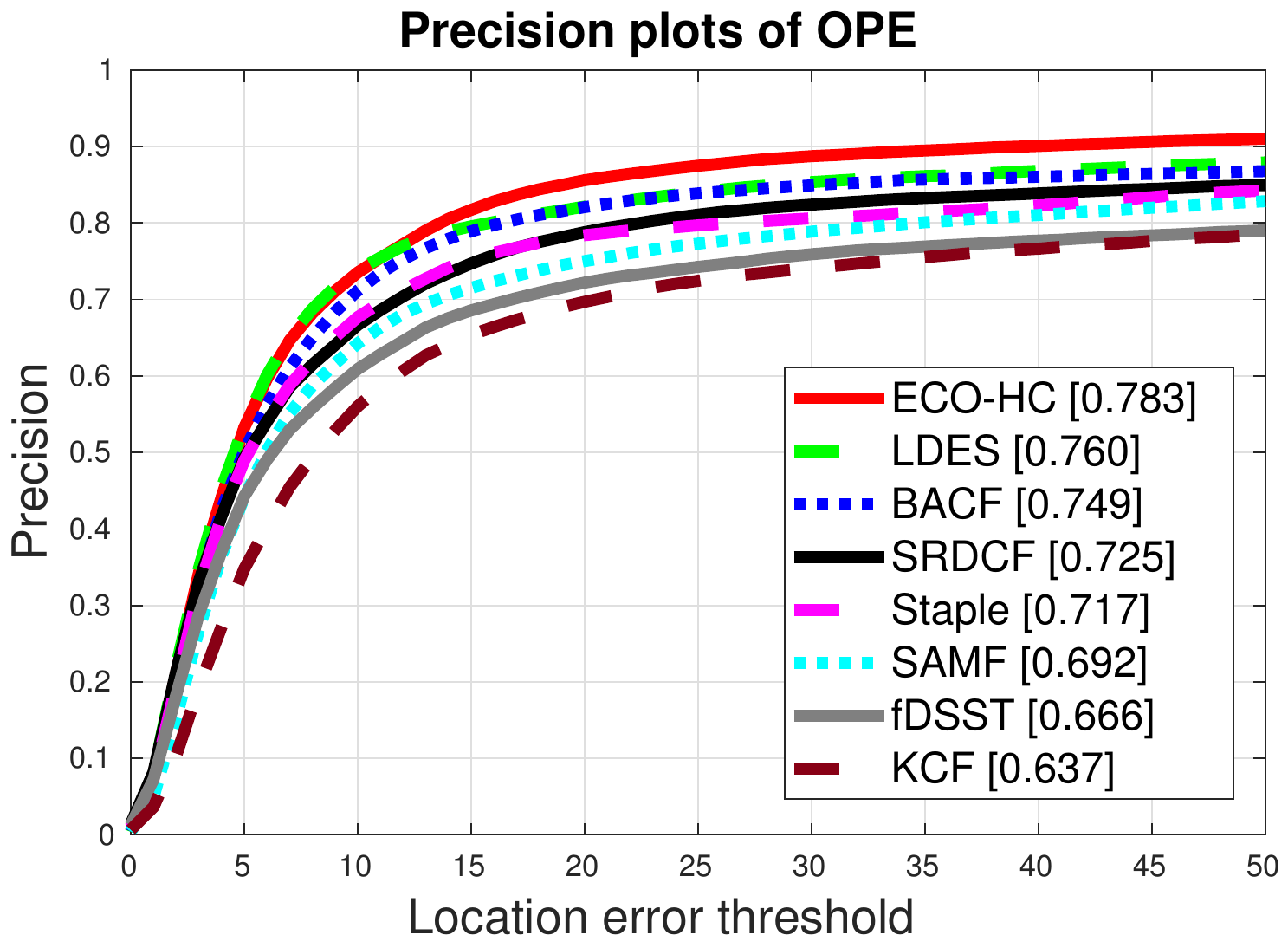}
}





\caption{The overall and scale performance in precision and success plots on OTB-2013 and OTB-100 dataset.}
\label{fig:tb100}
\end{figure*}

\section{Experiments}
In this section, we conduct four different experiments to evaluate our proposed tracker LDES comprehensively.

\subsection{Experimental Settings}

All the methods were implemented in Matlab and the experiments were conducted on a PC with an Intel i7-4770 3.40GHz CPU and 16GB RAM. 
We employ HoG feature for both translational and scale-rotation estimation, and the extra color histogram is used to estimate translational.
All patch is multiplied a Hann window as suggested in~\cite{bolme2010visual}.
$\eta$ is 0.15 and $\lambda$ is set to $1e^{-4}$. $\lambda_\phi$ and $\lambda_\alpha$ are both set to 0.01. $\lambda_\omega$ is 0.015.
The size of learning patch $D$ is 2.2 larger than the original target size. Moreover, the searching window size $N$ is about 1.5 larger than the learning patch size $D$. For scale-rotation estimation, the phase correlation sample size is about 1.8 larger than the original target size. All parameters are fixed in the following experiments.

\subsection{Experiments on Proposed Scale Estimator}
As one of the contributions in our work is a fast scale estimator, we first evaluate our proposed log-polar based scale estimator on OTB-2013 and OTB-100 dataset~\cite{wu2013online,wu2015object}. 
Three baseline trackers are involved in the scale estimation evaluation. They are SAMF~\cite{li2014scale},
fDSST~\cite{martin2017fdsst} and ECO~\cite{martin2017eco}. For fair comparison, we implement three counterpart-trackers including fDSST-LP, SAMF-LP and ECO-LP, which replace the original scale algorithm with our proposed scale estimator.

In Fig.~\ref{fig:compare}, these variant trackers with our scale component outperform their original implementation. This indicates that our proposed scale estimator has superior performance compared with current state-of-the-art scale estimator. Specifically, ECO-LP achieves 69.1\% and 67.3\% in OTB-2013 and OTB-2015 respectively, compared with its original CPU implementation's 67.8\% and 66.8\%. This proves the effectiveness of our proposed scale method since it can even improve the state-of-the-art tracker with a simple replacement of the scale component.

Since the proposed scale estimator only samples once in each frame, the most significant part is the efficiency of scale estimating. In Table~\ref{tab:speed}, the proposed approach has a 3.8X+ speedup on SAMF, and ECO, which obtains a significant improvement on efficiency. Even with fDSST which is designed in efficiency with many tricks, our method can still reduce its computational time. This strongly supports that our proposed scale estimator is superior to current state-of-the-art scale estimating approaches. 
In addition, our method is very easy to implement and plug-in to other trackers.

\def\lastfhtt{0.13\textheight}
\begin{figure} [!t]
  \normalsize
  \centering
  \includegraphics[height=\lastfhtt]{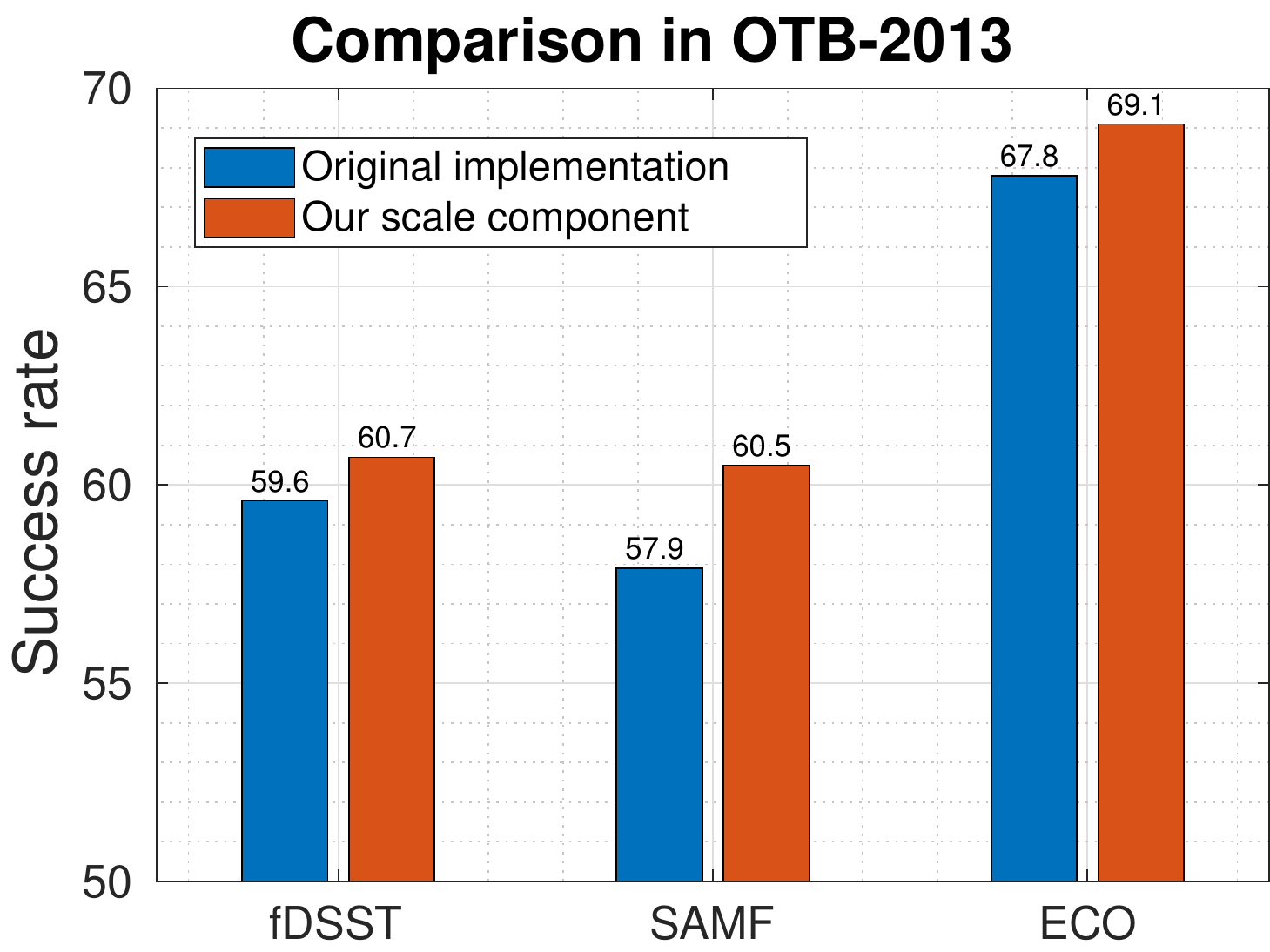}
  \includegraphics[height=\lastfhtt]{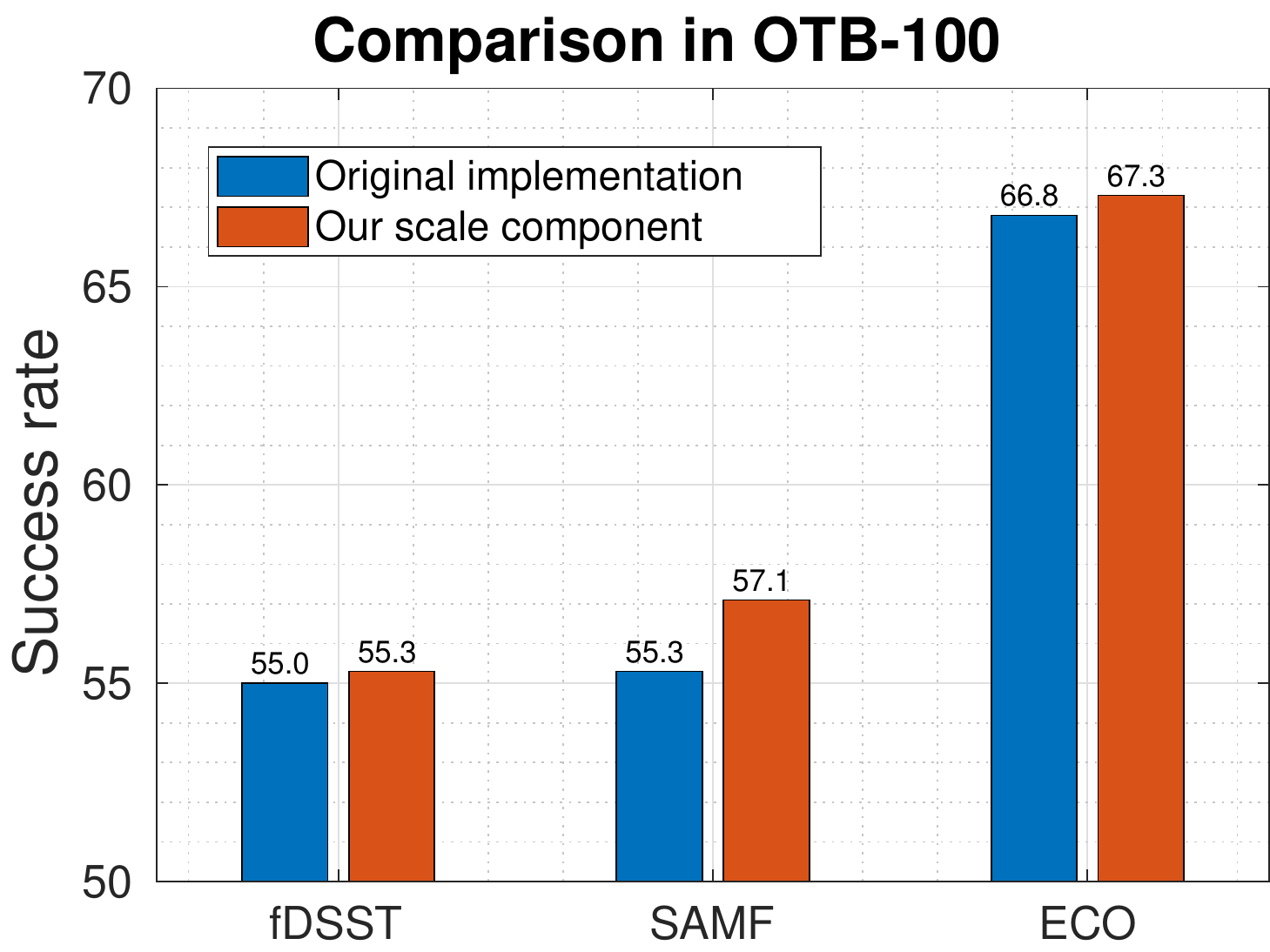}

     \caption{Evaluation of tracking success rate improvements over the original implementations of different trackers enhanced by the proposed scale estimator. }
     \label{fig:compare}

\end{figure}

\subsection{Comparison with Correlation Filter Trackers}
With efficient and effective scale estimator, our proposed tracker performs very promising in different situations.
We select seven state-of-the-art Correlation Filter-based trackers
as reference methods, including ECO-HC~\cite{martin2017eco}, SRDCF~\cite{martin2015srdcf},
Staple~\cite{bertinetto2015staple}, SAMF,
fDSST, 
BACF~\cite{galoogahi2017bacf}, and KCF~\cite{henriques2015high}.
We initialize the proposed tracker with axis-aligned bounding box and ignore the rotation parameter in the similarity transformation as tracking output since the benchmarks only provide axis-aligned labels.

In Fig.~\ref{fig:tb100}, 
it can be clearly seen that our proposed method outperforms most of the state-of-the-art correlation filter-based trackers and obtains 67.7\% and 81.0\% in OTB-2013 success and precision plots, and 63.4\% and 76.0\% in OTB-100 plots respectively.
ECO-HC achieves better results in OTB-100. However, we can see that our method is more accurate above 0.6 overlap threshold in success plot and comparable in the precision plot. The reason is that introducing rotation improves the accuracy but also enlarges the search space and hurts the robustness when large deformation occurs. In general, our method is very promising in generic object tracking task.

The proposed approach maintains 20 fps with similarity estimation and is easy to implement due to its simplicity. Moreover, our no-BCD version tracker achieves 82 fps in the benchmark while stall maintains comparable performance (67.5\% and 62.2\% accuracy in OTB-2013 and OTB-100, respectively). 

Please note that our proposed LDES is quite stable in searching the 4-DoF status space. Introducing rotation gives the tracker more status choice in tracking process while the benchmark only provides axis-aligned labels which make the performance less robust in OTB-100. However, our proposed tracker still ranks 1st and 2nd in OTB-2013 and OTB-100 respectively, and beats most of the other correlation filter-based trackers. 

\begin{table}[!]
  \centering
  \footnotesize
\caption{Evaluation of speedup results on different trackers achieved by applying the proposed Log-Polar~(LP) based scale estimator.}
\label{my-label}
\begin{tabular}{|l|c||l|c||c|}
  \hline
Trackers & FPS & Trackers & FPS & Speedup \\ 
\hline
fDSST & 101.31& fDSST-LP & 112.77  & 1.11X  \\  
\hline
SAMF &  20.95 &  SAMF-LP & 86.75 & 4.14X \\
\hline
ECO & 2.58 &  ECO-LP  & 9.88  & 3.82X\\  
\hline
\end{tabular}
\label{tab:speed}
\end{table}

\def\pot_h{0.13\textheight}
\begin{figure*}[!t]
  \centering
  \subfloat[All 210 sequences]
  {
    \includegraphics[height=\pot_h]{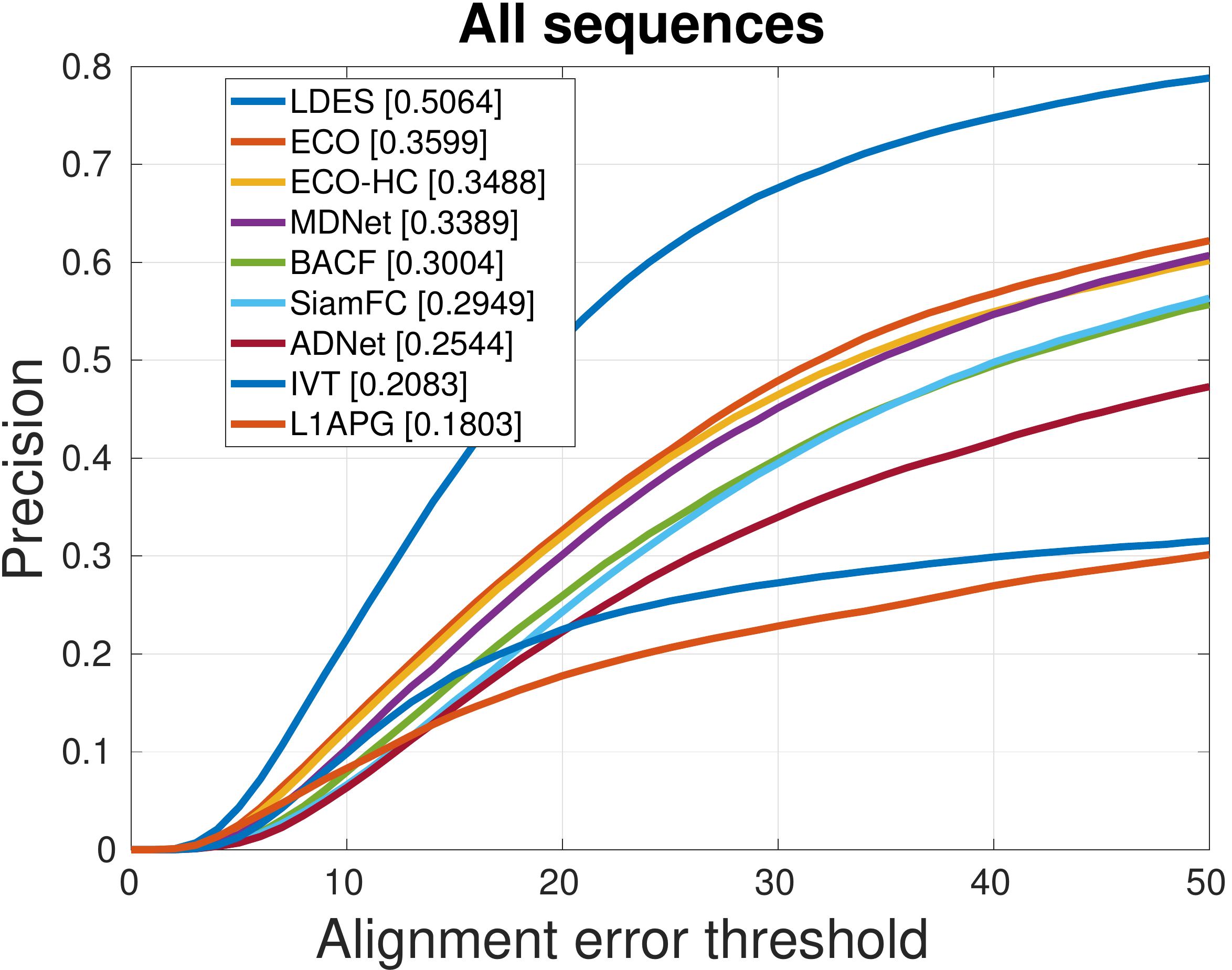}
    \includegraphics[height=\pot_h]{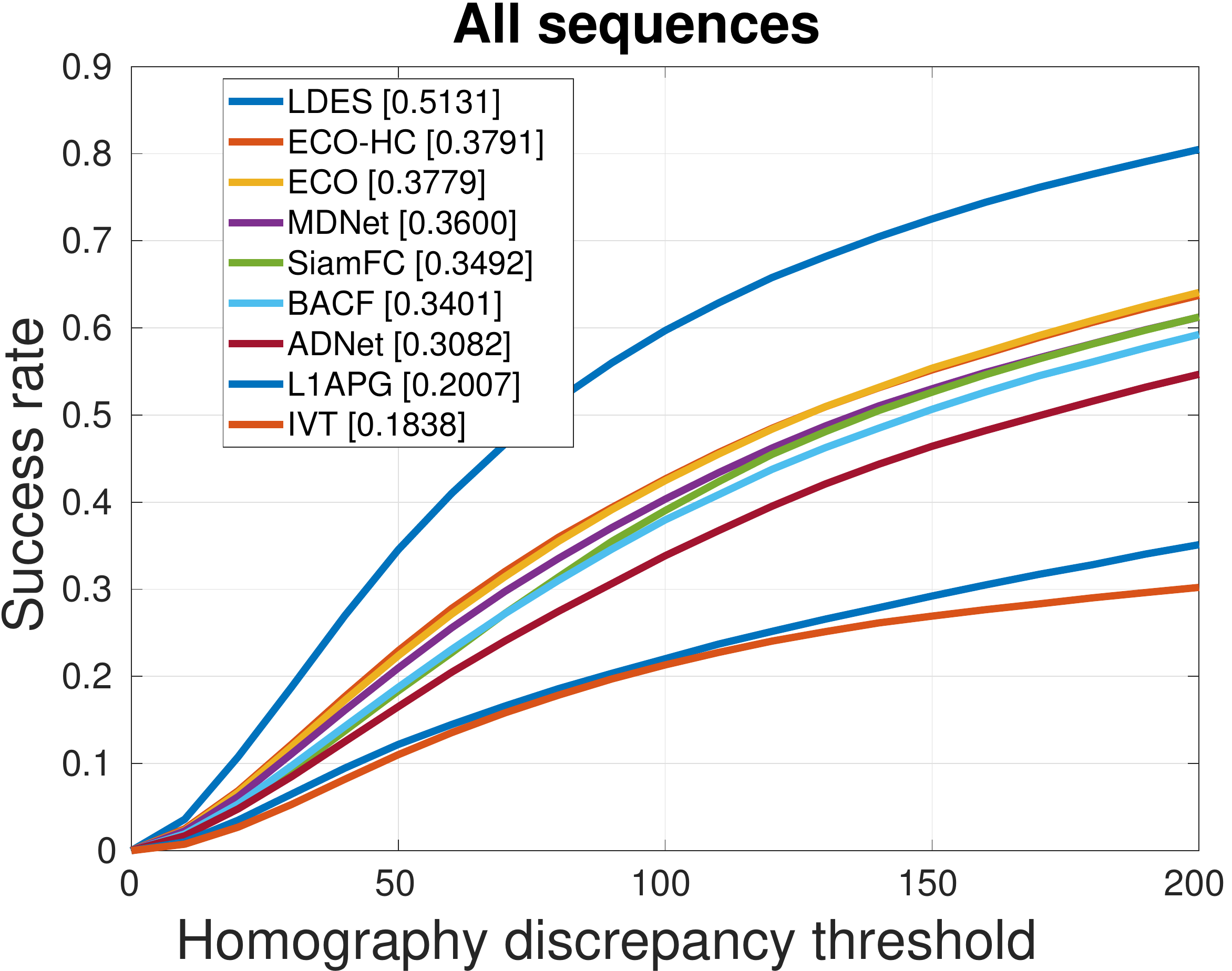}
    \label{fig:pot:a}
  }
  \subfloat[30 videos with unconstrained changes]
  {
    \includegraphics[height=\pot_h]{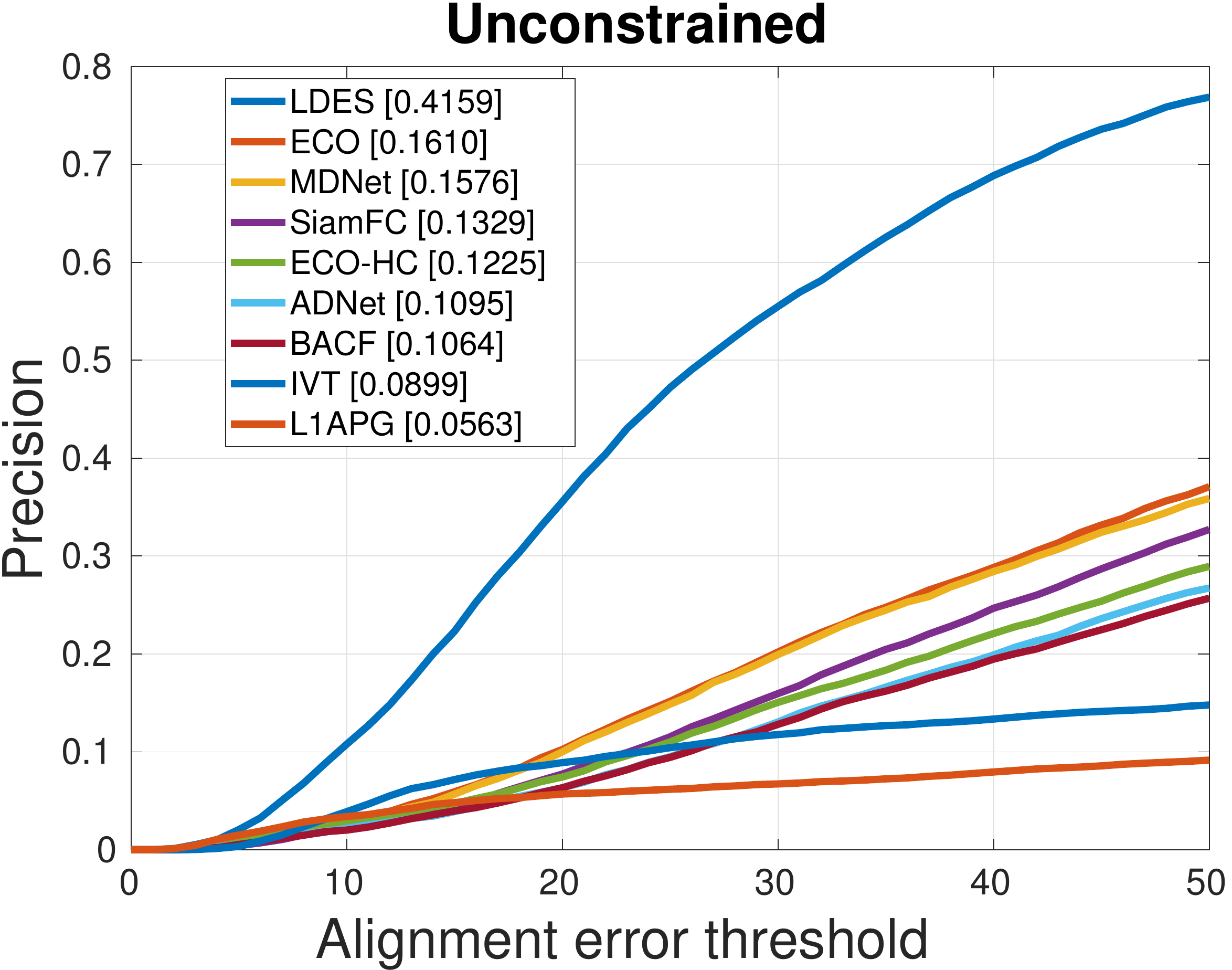}
    \includegraphics[height=\pot_h]{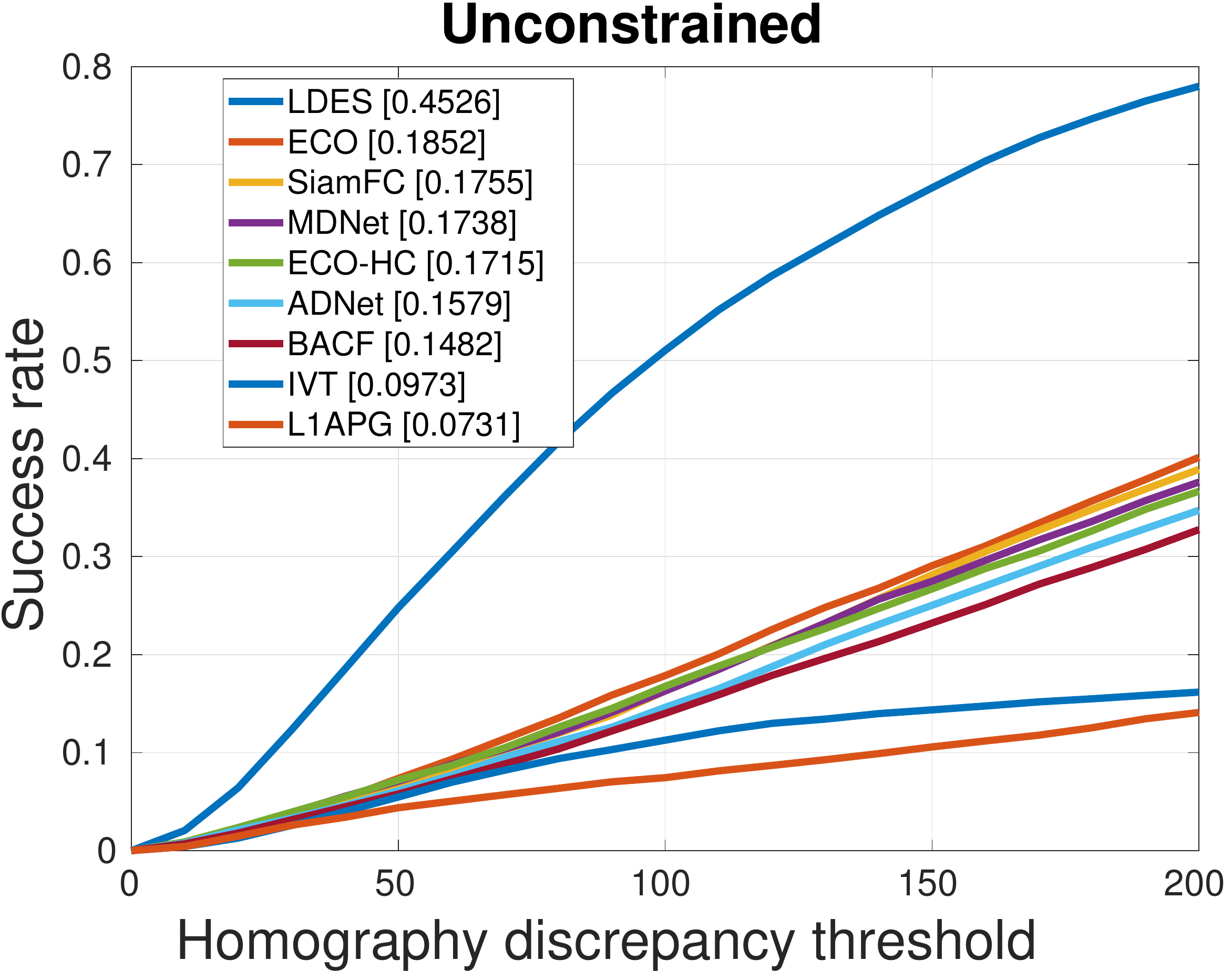}
    \label{fig:pot:b}
  }

  \subfloat[30 videos with pure rotation changes]
  {
    \includegraphics[height=\pot_h]{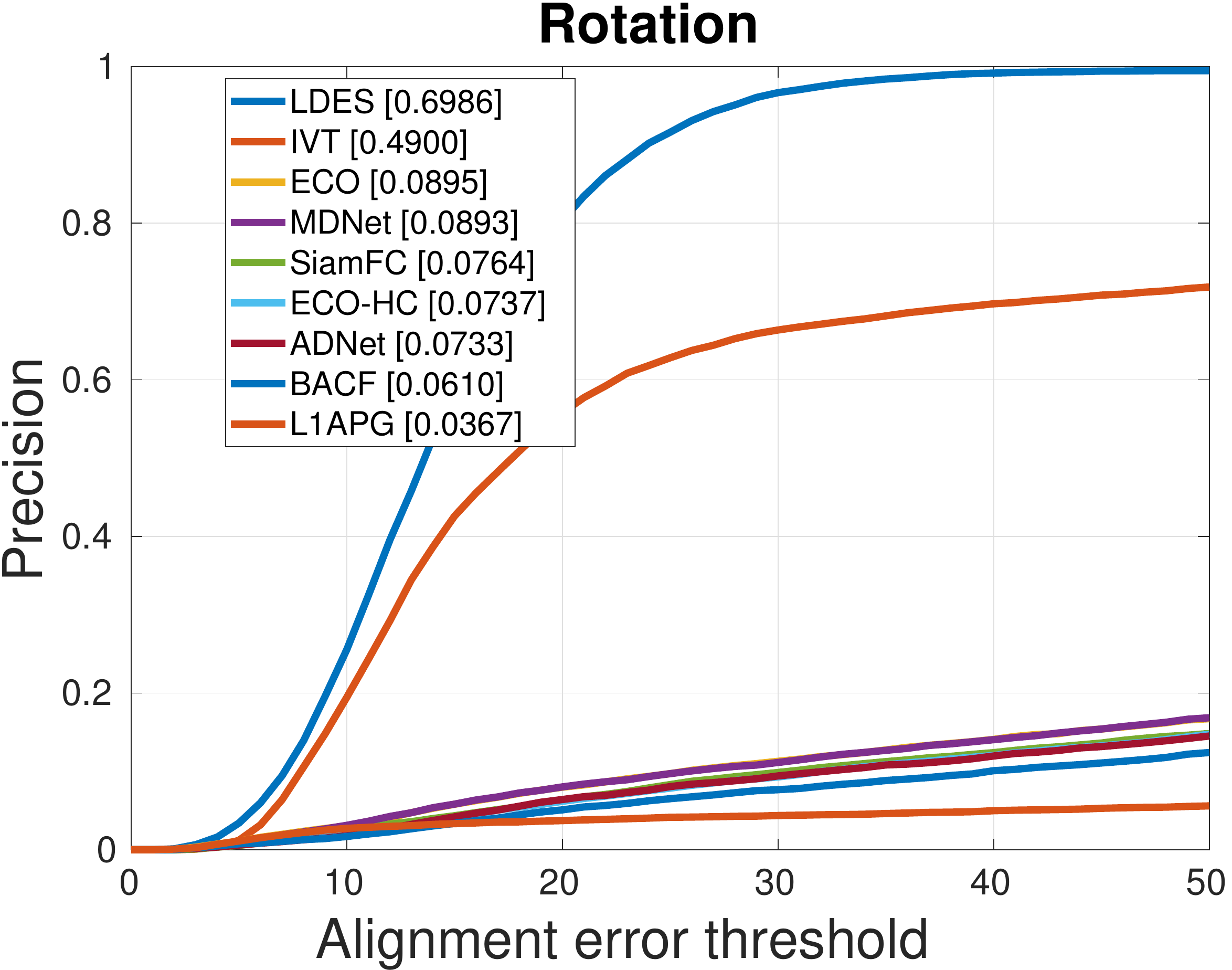}
    \includegraphics[height=\pot_h]{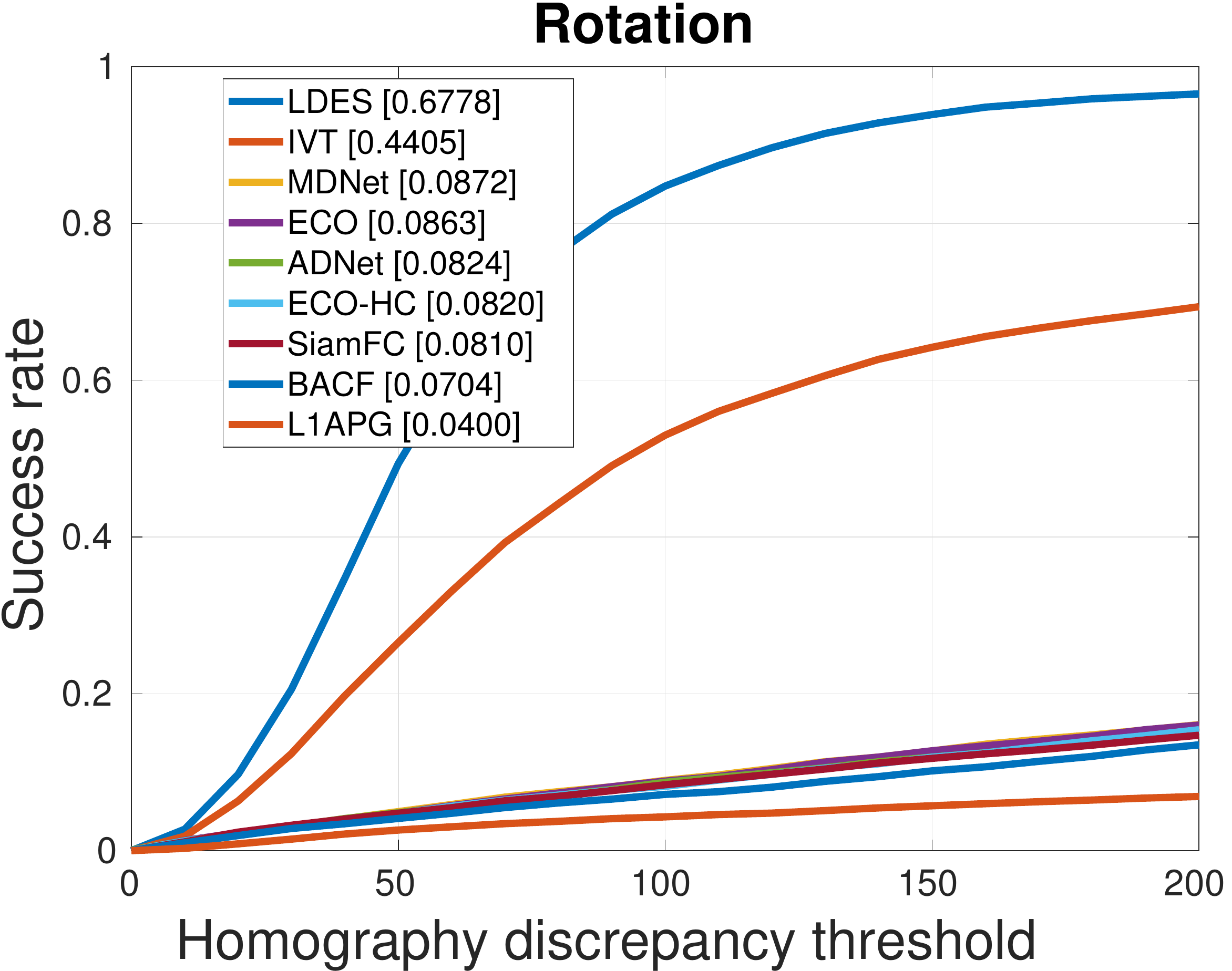}
    \label{fig:pot:c}
  }
  \subfloat[30 videos with pure scale changes]
  {
    \includegraphics[height=\pot_h]{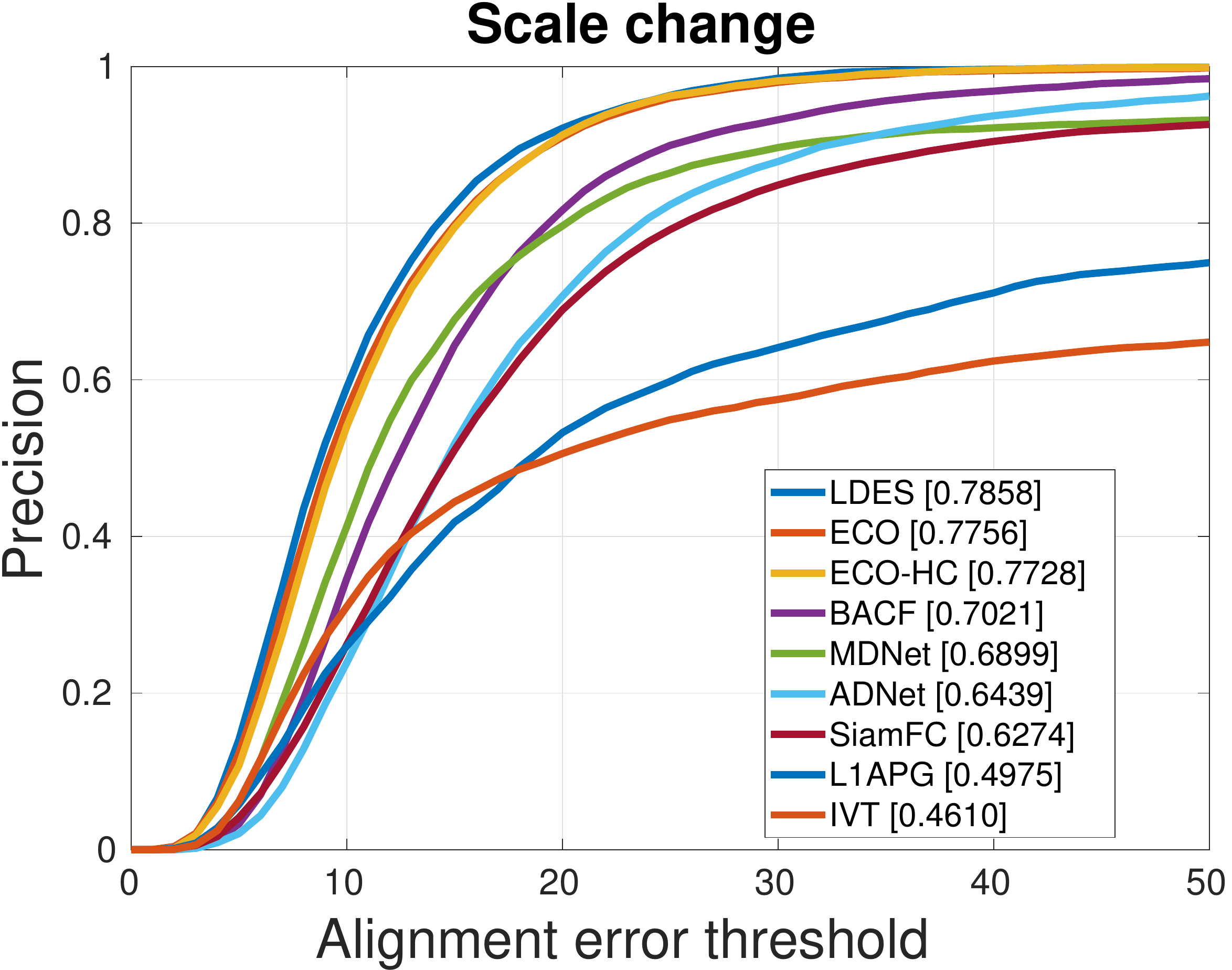}
    \includegraphics[height=\pot_h]{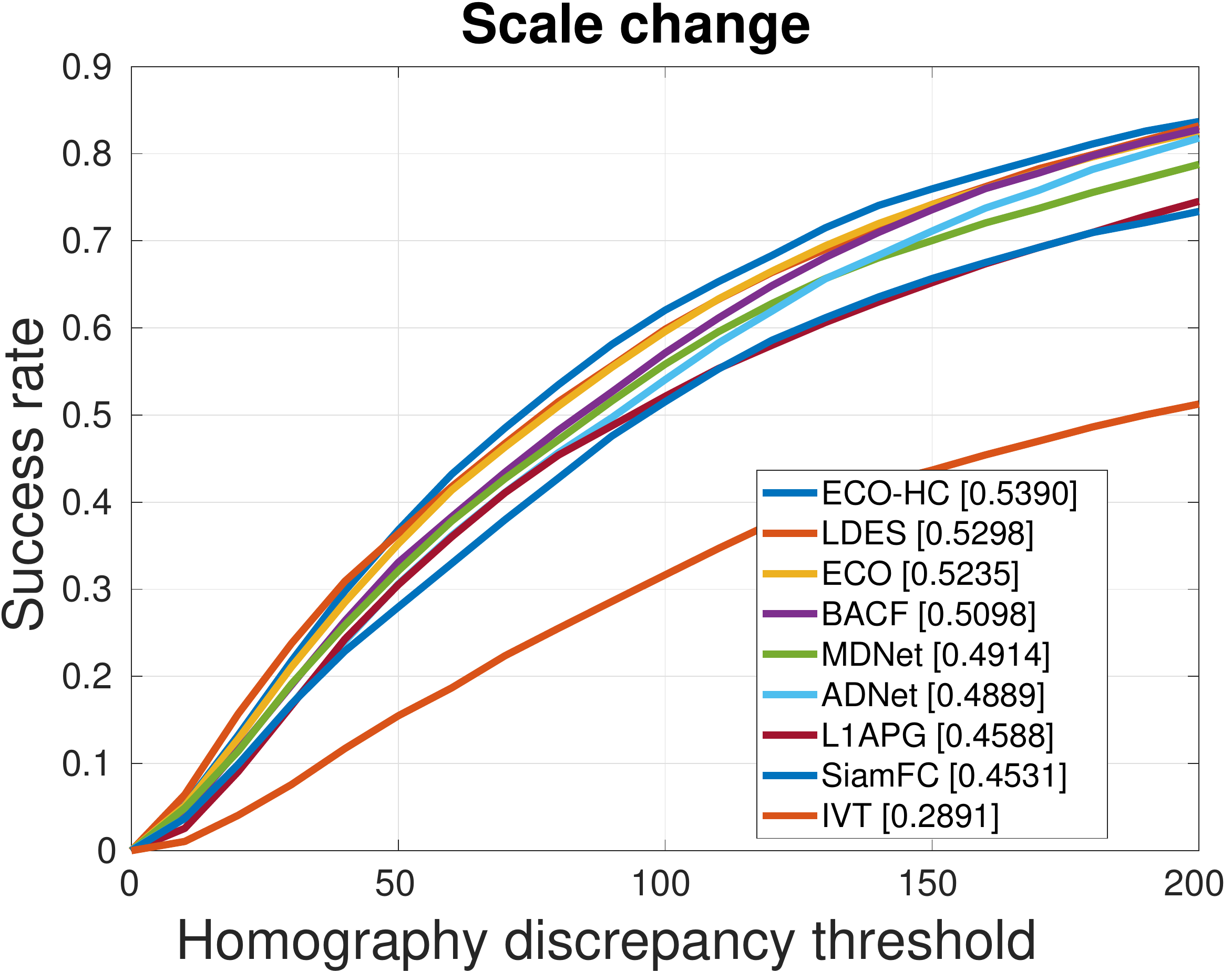}
    \label{fig:pot:d}
  }


\caption{Precision and success plots on the POT dataset.}

\label{fig:pot}
\end{figure*}

\subsection{Comparison with State-of-the-Art trackers on POT}

\begin{table*}[!t]
  \centering
  \footnotesize
\begin{tabular}{|c|c|c|c|c|c|c|c||c|}
  \hline
Precision 	&	Scale	&	Rotation	&	Persp. Dist.	&	Motion Blur	&	Occlusions	&	Out of View	&	Unconstrained	&	ALL\\ \hline
LDES		&	\textbf{0.7858}	&	\textbf{0.6986}	&	\textbf{0.2807}	&	\textbf{0.1699}	&	0.6277	&	\textbf{0.5663}	&	\textbf{0.4159}	&	\textbf{0.5064}\\ \hline
LDES-NoBCD  & 0.6461 & 0.6898 & 0.2528 & 0.1595 & \textbf{0.6679} & 0.5507 & 0.3759 & 0.4775\\ \hline
\hline
Success 	&	Scale	&	Rotation	&	Persp. Dist.	&	Motion Blur	&	Occlusions	&	Out of View	&	Unconstrained	&	ALL\\ \hline
LDES	&	\textbf{0.5298}	&	0.6778	&	\textbf{0.3339}	&	\textbf{0.44}	&	0.5947	&	\textbf{0.5629}	&	\textbf{0.4526}		&	\textbf{0.5131}\\ \hline
LDES-NoBCD  & 0.4724 & \textbf{0.6849} & 0.3215 & 0.43 & \textbf{0.6392} & 0.5615 & 0.4353 & 0.5064\\ \hline
\end{tabular}

  \caption{Comparison in all 7 categories attributed videos on POT benchmark.}
\label{tab:pot}
  \end{table*}



To better evaluate our proposed approach to rotation estimation, we conduct an additional experiment on POT benchmark~\cite{lian2017pot} which is designed to evaluate the planar transformation tracking methods.
The POT dataset contains 30 objects with 7 different categories which yield 210 videos in total.
Alignment error and homography discrepancy are employed as the evaluation metrics.
In addition, six state-of-the-art trackers and two rotation-enabled trackers are involved.
They are ECO-HC, ECO~\cite{martin2017eco}, MDNet~\cite{nam2015learning}, BACF~\cite{galoogahi2017bacf}, ADNet~\cite{yun2017adnet}, SiameseFC~\cite{BertinettoVHVT16}, IVT~\cite{ross2008incremental} and L1APG~\cite{Ji2012apga}.
To illustrate the POT plots appropriately, we set the maximal value of the alignment error axis from 20 to 50 pixels in precision plot and utilize the AUC as the metrics for ranking in both precision and homography discrepancy plots as same as OTB~\cite{wu2015object}. 

Fig.~\ref{fig:pot} shows that 
our proposed tracker, with hand-craft features only, performs extremely well in all sequences attributed plots and even outperforms deep learning based methods with a large margin.
In Fig.~\ref{fig:pot:a}, our LDES achieves 50.64\% and 51.31\% compared with second rank tracker ECO's 35.99\% and 37.79\% in precision and success rate plots within all 210 sequences, which is a 13\%+ performance improvement. 
Since the POT sequences are quite different from OTB, it indicates our proposed method has better generalization capabilities compared with pure deep learning based approaches in wide scenarios.
It also shows that our proposed method is able to search the 4-DoF similarity status simultaneously, efficiently and precisely.

Moreover, our method ranks 1st in almost all other plots.
It not only validates the effectiveness of our proposed rotation estimation but also shows the superiority of our method compared with traditional approaches.
In Fig.~\ref{fig:pot:d}, we argue that our proposed log-polar based scale estimation is at least comparable with mainstream methods in performance.


\subsection{BCD Framework Evaluation on POT}

To verify the proposed framework with Block Coordinate Descent (BCD), we implement an additional variant, named LDES-NoBCD, which turns off the BCD framework and only estimates the object status once in each frame. We conduct comparison experiments on POT benchmark with LDES and LDES-NoBCD.

In Table~\ref{tab:pot}, LDES performs better than its No BCD version in most of the categories. Specifically, BCD contributes more performance in scale attributed videos and unconstrained videos. LDES achieves 0.7858 and 0.5298 in scale compared with LDES-NoBCD's 0.6461 and 0.4724, which is about 14\% improvement in precision plot and 5\% in success plot, respectively. This indicates that the proposed framework ensures the stable searching in the 4-DoF space.

In rotation column, 
the ranks in precision and success rate metrics are inconsistent. The reason is that rotation attributed videos contain pure rotation changes. This gives rotation estimation a proper condition to achieve a promising result.
The only category that LDES performs inferior is occlusion attributed videos. When the occlusion occurs, BCD framework tries to find the best status of the templated object while the original object is being occluded and cannot be seen properly. This leads the algorithm to an inferior status. In contrast, No-BCD version algorithm does not search an optimal point in the similarity status space.






 \section{Conclusion}
In this paper, we proposed a novel visual object tracker for robust estimation of similarity transformation with correlation filter.  
We formulated the 4-DoF searching problem into two 2-DoF sub-problems and applied a Block Coordinates Descent solver to search in such a large 4-DoF space with real-time performance on a standard PC.
Specifically, we employed an efficient phase correlation scheme to deal with both scale and rotation changes simultaneously in log-polar coordinates and utilized a fast variant of correlation filter to predict the translational motion.
Experimental results demonstrated that the proposed tracker achieves very promising prediction performance compared with the state-of-the-art visual object tracking methods.

\section*{Acknowledgments}
This work is supported by the National Key Research and Development Program of China (No. 2016YFB1001501) and also by National Natural Science Foundation of China under Grants (61831015). This research is also supported by the National Research Foundation Singapore under its AI Singapore Programme [AISG-RP-2018-001], and the National Research Foundation, Prime Minister’s Office, Singapore under its International Research Centres in Singapore Funding Initiative.

\bibliography{bare_jrnl_compsoc}
\bibliographystyle{aaai}

\end{document}